\pdfoutput=1

\documentclass[11pt]{article}

\usepackage[final]{acl}

\usepackage{times}
\usepackage{latexsym}
\usepackage{booktabs}
\usepackage{multirow}
\usepackage{soul}
\usepackage{booktabs}
\usepackage{amssymb}
\usepackage{amsmath}
\usepackage[T1]{fontenc}

\usepackage[utf8]{inputenc}

\usepackage{microtype}

\usepackage{inconsolata}

\usepackage{tcolorbox}
\usepackage{algorithm}      
\usepackage{algpseudocode}  
\tcbuselibrary{skins} 
\tcbuselibrary{listings}
\definecolor{lightgreen}{RGB}{145, 203, 118}
\definecolor{lightyellow}{RGB}{250, 200, 88}
\definecolor{lightred}{RGB}{237, 101, 101}
\definecolor{lightblue}{RGB}{116, 193, 221}
\newtcolorbox{promptblock}[3][LLM Prompt]{
colback=black!5!white,
arc=3pt, 
boxrule=0.5pt,
fonttitle=\bfseries,
title=#1, 
before upper={\small}, fontupper=\fontfamily{ptm}\selectfont,
colframe=#2,
label=#3
}

\usepackage{makecell} 

\usepackage{pifont} 
\definecolor{darkgreen}{RGB}{0,160,0}
\newcommand{\notcheckmark}{\textcolor{black}{\bcmark\kern-1.1ex\raisebox{.7ex}{\rotatebox[origin=c]{125}{--}}}\color{black}}
\newcommand{\bcmark}{\color{blue}{\ding{51}}}%
\newcommand{\cmark}{\color{darkgreen}{\ding{51}}}%
\newcommand{\xmark}{\color{red}{\ding{55}}}%

\usepackage{graphicx}
\title{ChartM$^3$: A Multi-Stage Code-Driven Pipeline for Constructing Multi-Dimensional and Multi-Step Visual Reasoning Data in Chart Comprehension}

\author{Duo Xu\thanks{The first two authors contributed equally}, Hao Cheng\footnotemark[1], Xin Lin, Zhen Xie \& Hao Wang\protect \thanks{ Corresponding author}
\\
Alibaba Cloud Computing
\\
\tt {manii.xd@alibaba-inc.com, haochworktime@gmail.com, cashenry@126.com}
}

\begin{document}
\maketitle
\begin{abstract}

Complex chart understanding tasks demand advanced visual recognition and reasoning capabilities from multimodal large language models (MLLMs). However, current research provides limited coverage of complex chart scenarios and computation-intensive reasoning tasks prevalent in real-world applications. This study proposes an automated multi-stage \textit{code-driven} pipeline for systematically generating visual reasoning datasets to address these limitations. The pipeline integrates retrieval-augmented generation (RAG) to retrieve professional chart templates and employs chain-of-thought (CoT) strategies to generate reasoning codes that simulate real data distributions, thereby driving chart rendering and question-related statistical computations. Through model-based evaluation, the pipeline enhances chart diversity and data quality. Using this framework, we construct \textbf{ChartM$^3$}, a multi-dimensional and multi-step dataset containing 38K charts and 142K Q\&A pairs for training, along with 2,871 high-quality evaluation samples for enabling practical performance assessment. Supervised fine-tuning (SFT) and reinforcement learning (RL) experiments demonstrate that our dataset significantly improves reasoning capabilities and cross-domain generalization performance, enabling smaller models to achieve performance comparable to larger-scale models in complex chart comprehension. 

\end{abstract}

\section{Introduction}
\label{sec:introduction}

Advanced Multimodal Large Language Models (MLLMs) such as GPT-4o \cite{jaech2024openai}, LLaVA \cite{liu2023llava}, Qwen-VL \cite{Qwen2.5-VL, Qwen-VL}, and InternVL \cite{chen2024internvl} series have continuously emerged, demonstrating remarkable capabilities in Visual Question Answering (VQA) for natural images. However, these models still struggle with text-rich images, particularly in chart comprehension. Unlike natural images, which primarily focus on perceptual understanding, charts are intricate visual systems that combine multiple elements (titles, legends, axes, etc.) to present structured data. Effectively understanding charts requires processing visual information, analyzing the hierarchical relationships between these elements, and interpreting the underlying design intent.

\begin{figure*}[t]
\centering
  \includegraphics[width=0.40\linewidth]{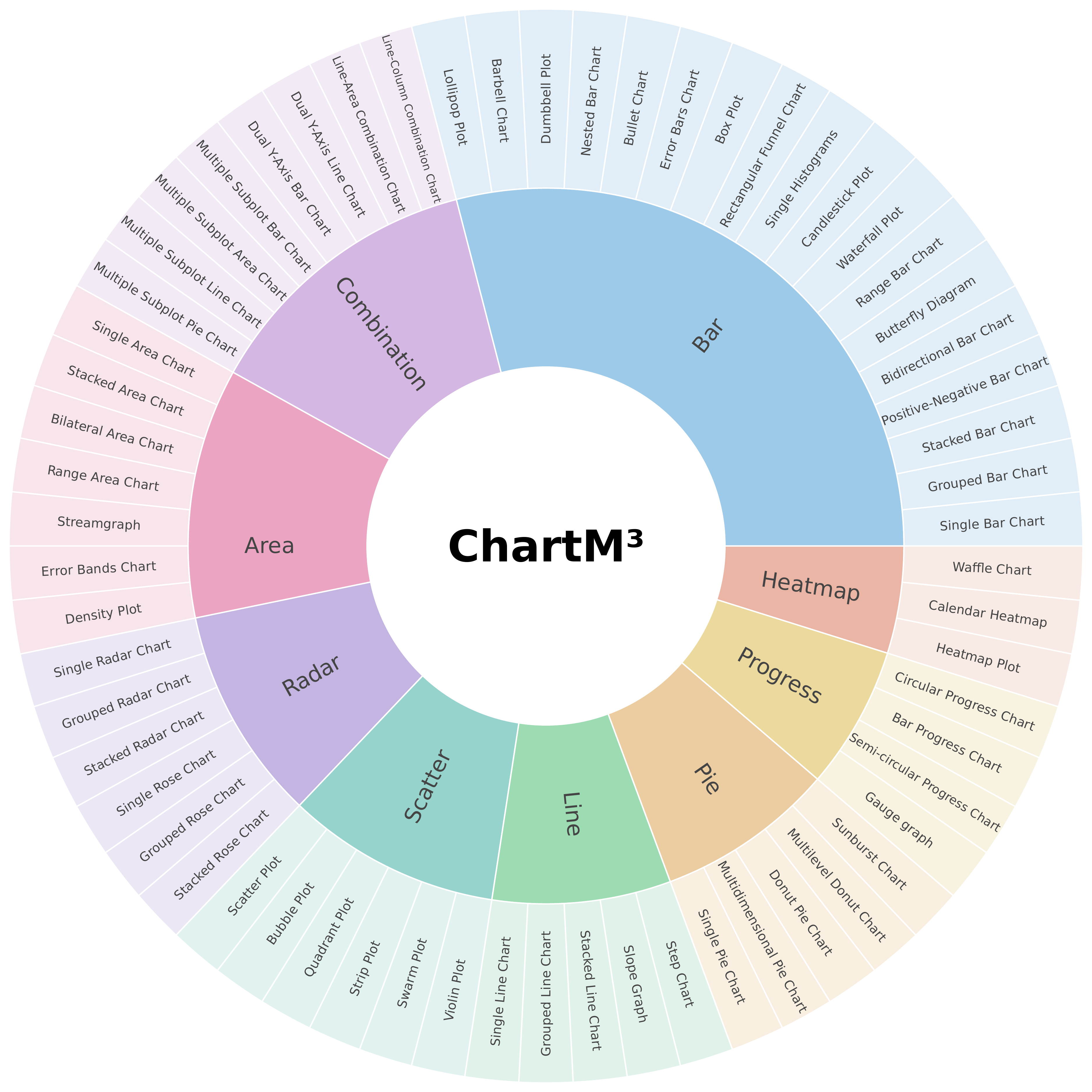} \hspace{0.05\linewidth}
  \includegraphics[width=0.45\linewidth]{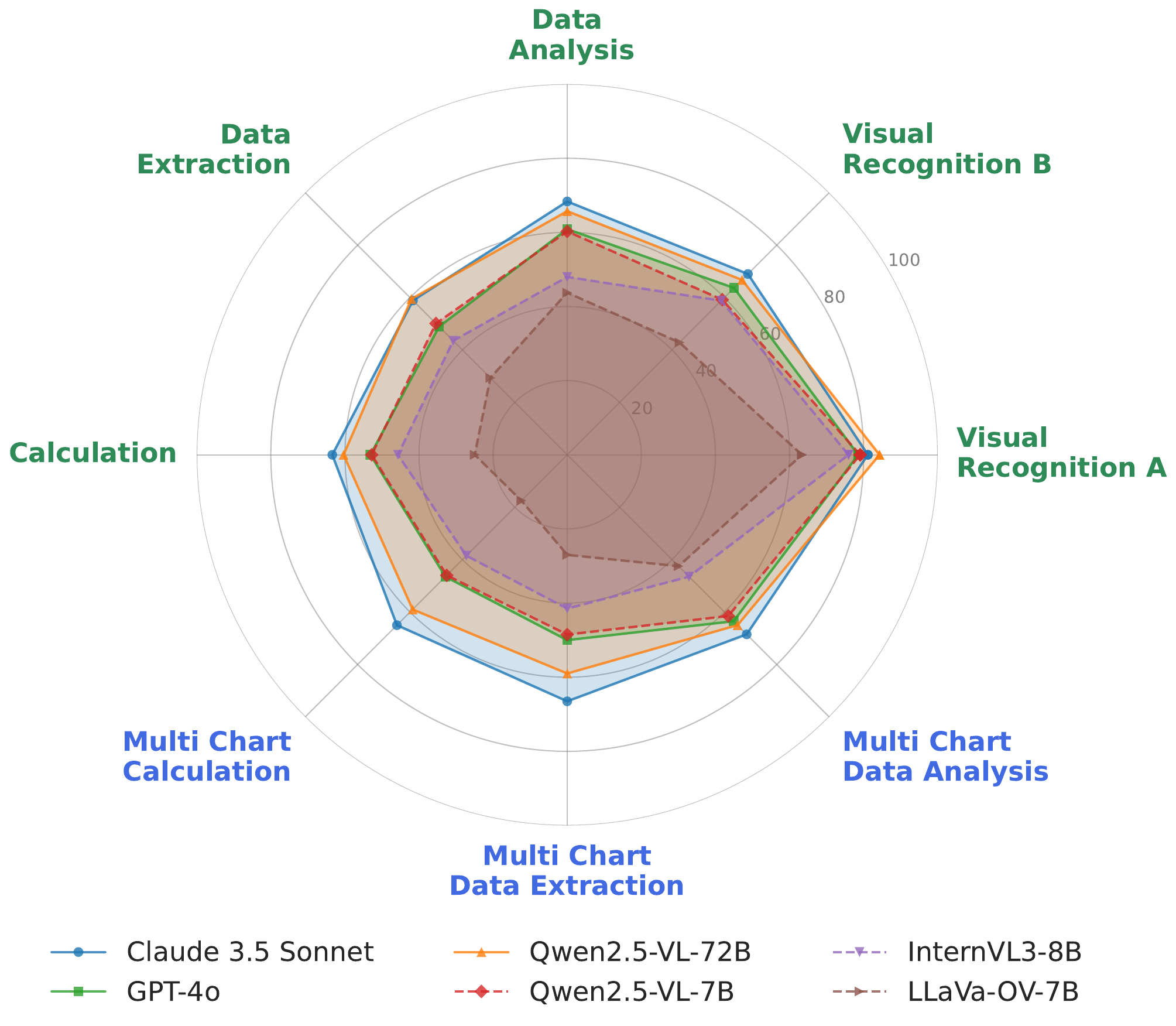}
  \caption {Left: ChartM$^3$ covers 9 major categories of chart types, totaling 62 subcategories. Right: Performance comparison of representative MLLMs across ChartM$^3$ task categories.}
  \label{fig:ChartM3}
\end{figure*}

Despite strong benchmark performance on ChartQA \cite{masry-etal-2022-chartqa} and PlotQA \cite{Methani_2020_WACV}, state-of-the-art MLLMs often deliver unsatisfactory results in real-world applications. This discrepancy arises from the complexity of actual charts, which significantly exceeds that of benchmark datasets. Current chart datasets \cite{xia2024chartx, ChartBench} exhibit several critical limitations:  \textbf{Limited Chart Type and Element Complexity.} Most existing datasets primarily focus on compositionally simple charts, such as line, bar, and pie charts, while neglecting data-intensive formats like scatter plots and heatmaps, or sophisticated derivatives such as box plots and multi-axis composites.  \textbf{Low Question Complexity.}  Current datasets emphasize basic perceptual tasks rather than complex business analytics that demand multi-step reasoning and multi-chart comprehension. \textbf{Lack of Interpretability Support.} These datasets focus solely on question-answer pairs without providing detailed stepwise reasoning processes to enhance model understanding, limiting data utility and model explainability in practical applications. These limitations originate from inherent conflicts between data accuracy, complexity, and construction costs in conventional data creation approaches.

To address these challenges, we introduce \textbf{ChartM$^3$}, a comprehensive chart dataset that extends both chart variety and task complexity while reflecting real-world analytics scenarios. 
Our automated pipeline decomposes the generation process into a four-stage chain: database construction, data code generation, visualization code creation, and Q\&A pair synthesis with reasoning code. Each stage is implemented through executable Python code to ensure traceability and verifiability. 
The process begins by constructing a diverse chart template database including 62 chart types and generates high-quality questions across 60 real-world scenarios. Using Retrieval-Augmented Generation (RAG) to select professional templates, we employ LLM's Long Chain-of-Thought (CoT) reasoning to thoroughly analyze data generation context and visualization requirements. This CoT-driven approach generates both structured data and visualization code, followed by MLLMs formulating questions and synthesizing analytical code with reliable reasoning paths. Through code execution and output verification, we produce accurate answers with reliable CoT reasoning.
To further enhance quality, we employ a combination of large and small language models to filter out unsuitable charts and Q\&A pairs. This Multi-stage, Multi-dimensional, and Multi-step (M$^3$) approach guarantees data quality and diversity, resulting in a comprehensive dataset containing 38.4K diverse charts and 142K high-quality Q\&A pairs, and a challenging benchmark of 2,871 rigorously verified samples.

We validate the effectiveness of ChartM$^3$ through comprehensive experiments, demonstrating substantial improvements in business insight extraction and analytical reasoning capabilities. This dataset advances the development of practical chart understanding systems and helps bridge the gap between academic evaluation and real-world applications.

Our contributions can be summarized as follows:
\begin{itemize}

    \item We present a novel pipeline that leverages open-source LLMs to synthesize aligned chart data and visual reasoning Q\&A pairs. Through RAG for template retrieval, code-driven generation, and model-based quality control, our approach produces diverse, professional-quality synthetic chart data.
    
    \item We construct a comprehensive benchmark that systematically identifies architectural limitations in complex chart comprehension and cross-chart reasoning capabilities.

    \item Comprehensive experiments demonstrate that models trained on ChartM$^3$ show substantial improvements in visual perception and reasoning abilities, validating that our framework provides a practical methodology for developing reasoning MLLMs.
    
\end{itemize}

\section{Related Works}
\label{sec:related-works}



\begin{table*}[htbp]
  \footnotesize
  \centering
    \begin{tabular}{lccccccc}
    \toprule
    \multirow{3}{*}{\textbf{Datasets}} & \multicolumn{3}{c}{\textbf{Chart Properties}} & \multicolumn{4}{c}{\textbf{Q\&A Properties}} \\
    \cmidrule(lr){2-4} \cmidrule(lr){5-8} 
    & \makecell{Data Source} & \makecell{\# Chart\\Type} & \makecell{Textual\\Data} & \makecell{\# Task\\Type} & \makecell{Template-Free\\Question} & \makecell{Multi Chart\\Q\&A} & \makecell{Reasoning\\Data} \\
    \midrule
    FigureQA & Synthetic & 5 & - & 15 & \xmark & \xmark & \xmark \\
    DVQA & Synthetic & 1 & - & 3 & \xmark & \xmark & \xmark \\
    PlotQA & Real-world, Synthetic & 4 & Table & 3 & \xmark & \xmark & \xmark \\
    ChartQA & Real-world, Synthetic & 3 & Table & 4 & \cmark & \xmark & \xmark \\
    ChartLLama & Synthetic & 10 & Table & 7 & \cmark & \xmark & \xmark \\
    MMC-Instruction & Real-world & 6 & Caption & 9 & \cmark & \cmark & \cmark \\
    ChartBench & Real-world, Synthetic & 42 & Table & 5 & \cmark & \cmark & \xmark \\
    ChartX & Synthetic & 18 & Code & 7 & \cmark & \xmark & \xmark \\
    OneChart & Real-world, Synthetic & 7 & Table & 1 & \xmark & \xmark & \xmark \\
    ChartAst (ChartSFT) & Real-world, Synthetic & 9 & Table & 5 & \cmark & \xmark & \cmark \\
    ChartInstruct & Real-world, Synthetic & 13 & - & 6 & \cmark & \xmark & \cmark \\
    CharXiv & Real-world & - & - & 23 & \xmark & \cmark & \xmark \\
    ChartGemma & Real-world, Synthetic & - & Caption & 10 & \cmark & \xmark & \cmark \\
    MultiChartQA & Real-world & - & - & 4 & \cmark & \cmark & \xmark \\
    ReachQA & Synthetic & 32 & Code & 3 & \cmark & \cmark & \cmark \\
    \textbf{ChartM$^3$(Ours)} & {\textbf{Synthetic}} & {\textbf{62}} & \textbf{Code} & {\textbf{18}} & \cmark & \cmark & \cmark \\
    \bottomrule
    \end{tabular}%
  \label{tab:chart-benchmarks}%
\caption{Comparison of Several Datasets for Chart QA.}
\end{table*}%

For chart comprehension and question-answering datasets, early studies (such as FigureQA~\cite{kahou2017figureqa}, DVQA~\cite{kafle2018dvqa}) proposed synthetic data generation pipelines to produce VQA datasets for several chart types (typically 1-4 types of charts).
However, these approaches were constrained by the limitations of the synthetic data pipelines at the time, resulting in issues such as limited chart topics, templated task types, and fixed answer formats. 
PlotQA \cite{Methani_2020_WACV} expanded the range of chart topics by introducing real-world data but focused only on bar charts, line graphs, and scatter plots. Moreover, the program-synthesized charts had relatively simple styles, with visual designs and color schemes that could hardly represent real-world standards. 
ChartQA \cite{masry-etal-2022-chartqa} further broadened the scope of question forms and openness through human annotation and machine generation, breaking free from template-based restrictions on questions. Nevertheless, it still suffered from a limited variety of chart types. 
MMC-Instruction \cite{liu2023mmc}, ChartBench \cite{ChartBench}, and CharXiv \cite{wang2024charxiv} improved the diversity of chart and question types by collecting real-world chart data and combining them with manual annotations, but this also led to increased costs and limited scalability. 

In recent years, with the continuous advancement of large language models (LLM), researches have been utilizing LLMs for data synthesis have emerged. Compared to template-based data synthesis pipelines, these works have significantly improved chart topic richness and Q\&A flexibility. 
For example, ChartLlama \cite{han2023chartllama}, ChartInstruct \cite{masry2024chartinstruct}, and TinyChart \cite{zhang2024tinychartefficientchartunderstanding} generate data, plotting code, and Q\&As through pipelines. Research like ChartAssistant (ChartSFT) \cite{meng2024chartassisstantuniversalchartmultimodal} and ChartGemma \cite{masry2024chartgemma} utilizes existing synthetic and real-world datasets to construct instruction datasets for chart understanding model training.
However, there is still room for improvement in fine-grained visual element analysis (e.g., layout, color style). 
Regarding evaluation tasks, ChartInsights \cite{wu2024chartinsights} systematically defines structural parsing tasks for seven types of charts, revealing deficiencies in mainstream models like GPT-4V in low-level tasks such as axis recognition and legend matching (with an average accuracy below 60\%). 
ChartX \cite{xia2024chartx} further extends the evaluation dimensions by supporting seven subtasks, including structure extraction and cross-modal generation, with 48k quadruples (image-CSV-code-text). 
However, current chart datasets still face challenges in constructing complex scenario questions and multi-step reasoning tasks, with evaluation pipelines that are not sufficiently objective. As a result, existing datasets still cannot accurately measure the true chart comprehension capabilities of MLLMs. In this article, we introduce ChartM$^3$, a novel chart dataset produced by an automatic multi-stage data synthesis pipeline designed for high-quality visual reasoning chart Q\&A data.

\section{ChartM$^3$}
\label{sec:chartm3}

\begin{figure*}[t]
  \centering
  \includegraphics[trim=0mm 0mm 0 0, clip, width=1\linewidth]{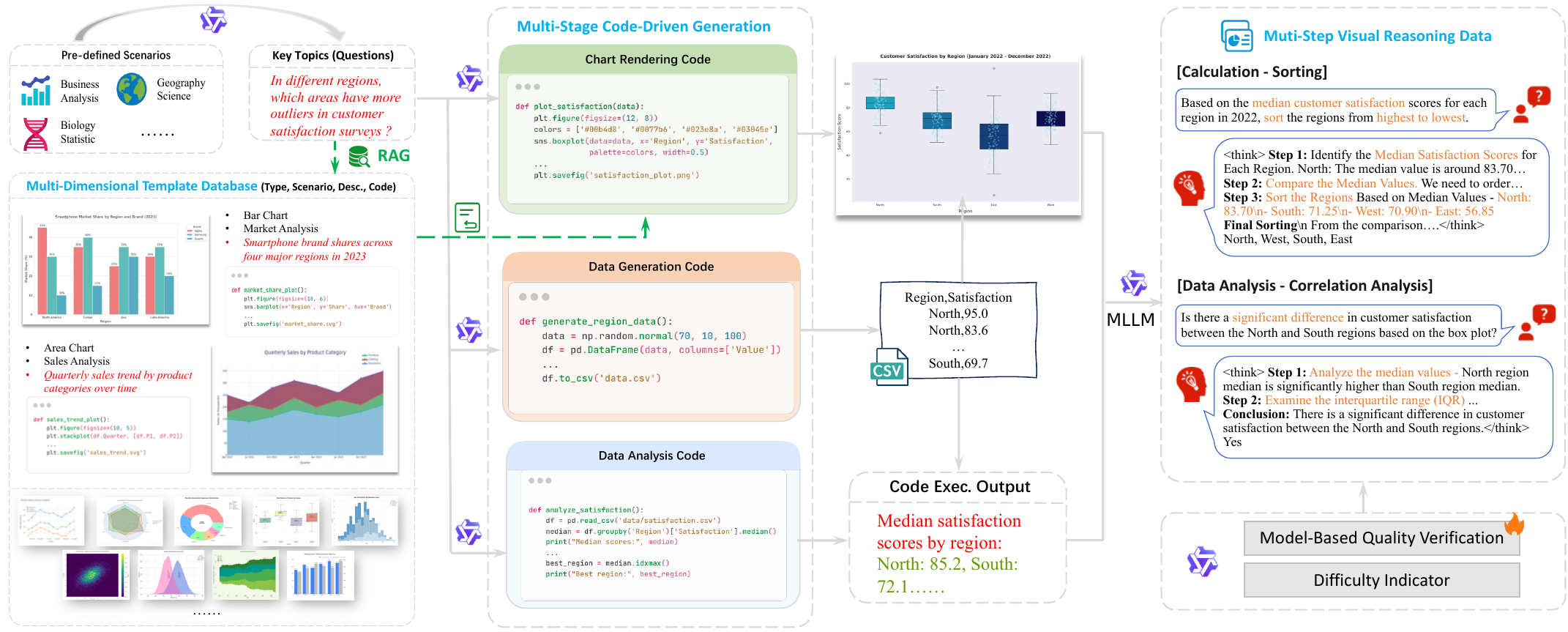}
  \caption{The ChartM$^3$ data generation pipeline follows a progressive automated workflow that begins by generating key questions and utilizing RAG to select appropriate templates from a diverse chart database. The process then advances through multiple code-driven stages: creating structured data, producing rendering code, and generating Q\&A pairs with multi-step visual reasoning reasoning synthesizing analytical code. Finally the pipeline conducts model-based comprehensive assessments of data quality and difficulty levels.}
  \label{fig:pipeline}
\end{figure*}

Figure~\ref{fig:pipeline} illustrates the ChartM$^3$ automated workflow. Our core approach combines RAG-based chart template selection with a multi-stage, code-driven generation process and model-based quality verification. Beyond single-chart analysis, we also incorporate cross-chart comparison tasks that require examining multiple images simultaneously. The following sections detail each stage of implementation: template database construction (\textsection~\ref{sec:31}), chart data and image generation (\textsection~\ref{sec:32}), instructional Q\&A generation (\textsection~\ref{sec:33}), and data evaluation (\textsection~\ref{sec:34}). Based on our dataset, we introduce a novel reinforcement learning approach for chart comprehension tasks, as detailed in (\textsection~\ref{sec:35}).

\subsection{Template Database Construction} \label{sec:31}

We develop a comprehensive chart taxonomy by analyzing major visualization frameworks such as Matplotlib \cite{Hunter:2007}, Vega \cite{2017-vega-lite}, EChart \cite{LI2018136}, and Seaborn \cite{michael_waskom_2017_883859}. Our analysis identifies 62 scientifically rigorous chart types commonly used in real-world scenarios (shown in Figure~\ref{fig:ChartM3}). Each type of chart is annotated with descriptive tags covering definitions, usage scenarios, and data characteristics. 

For Database generation, we utilize Claude 3.5 to create structured data and code templates for each chart type, incorporating comprehensive parameters for standardized rendering. To enhance visualization diversity, we develop templates that align with real-world scenarios across themes, layouts, and color schemes. We incorporate domain-specific styles from various professional fields and manually refine the details to better align with real-world charts. In addition, we collect real-world charts from various sectors, including finance and scientific research. These charts are recreated using Claude 3.5 to generate style-matching code templates. Each chart template is labeled with multiple attributes, including industry domain, theme, and visualization purpose, all constructed based on visual characteristics and type descriptions.

\subsection{Chart Image Generation} \label{sec:32}
Instead of direct data generation, we divide this building process into multiple substages with a code-driven method to avoid distributional convergence in LLM-generated content. We curate 60 domains commonly associated with data visualization and create key questions that require analytical reasoning rather than generating random titles. This approach reflects the purpose-driven nature of real-world charts, typically designed to address specific problems or analyze trends. Using the domain and questions as input, we leverage RAG to dynamically match the most representative chart types and suitable templates from the template database.

LLMs then transform these key questions into realistic contextual narratives and develop corresponding structured data and metadata (including titles and descriptions). To prevent distributional monotony and errors in large-scale data generation, we require LLMs to output data generation code rather than direct data. LLMs are prompted to incorporate data distribution trends, stochastic functions, and controlled noise into their code. 

During the generation of visualization code, we use a step-by-step reasoning approach to enhance code usability and visual quality. The process begins by guiding LLMs through visualization requirement analysis, which includes evaluating data and industry background and developing a detailed solution of visual elements. To increase visual diversity, we randomly integrate style-enhancing prompts during this phase. Using the generated visualization solution and selected template code as few-shot demonstrations, we produce and execute visualization code to generate chart images. If code execution fails, we feed the code and error messages back to LLMs for iterative refinement.

\subsection{Instruction Q\&A Generation} \label{sec:33}

We develop 18 specialized Q\&A categories across four primary dimensions based on perception and reasoning levels: visual element recognition, data extraction, calculation, and data analysis. These tasks span multiple formats (Multiple-choice, True/False, Fill-in-the-blank, and Short-answer) and are designed to elicit in-depth thinking and multi-step reasoning. Using visualization code, data, and task specifications as inputs, we guide LLMs to systematically generate questions through carefully crafted prompts and ICL examples from real-world scenarios or other datasets, as detailed in Appendix~\ref{appendix:prompt_templates}.

Our approach identifies two critical challenges in LLM-synthesized data: (1) potential information misalignment between plotting code and rendered images in complex charts, and (2) high error rates in numerical comparison and complex computation tasks from open-source models. 
To address these, we leverage Qwen2.5-VL-72B to focus exclusively on visual information during question generation, while adopting an agent-inspired approach for computational tasks. This approach generates executable code snippets for problem-solving, using the execution outputs and intermediate steps to construct answer and comprehensive reasoning paths.


\subsection{Data Evaluation} \label{sec:34}

Since we heavily depend on LLM synthesis throughout the process, building on the basic filtering of abnormal outputs and code execution failures, we implement several quality control modules which employ multiple models collaboratively for multi-dimensional quality assessment:

\noindent\textbf{Chart Quality Verification.} Our experiments reveal that even MLLMs with up to 72B parameters struggle to reliably evaluate chart quality, often missing issues like data occlusion or suboptimal layout arrangements. Using MLLMs pre-labeling as a starting point, we correct erroneous results to create a chart quality classification dataset comprising 700 positive and 500 negative samples. We then train a classifier based on Qwen2-VL-2B, which achieve a higher F1 score on the validation set compared to Qwen2.5-VL-72B.

\noindent\textbf{Instruction Verification.} We implement a multi-modal verification step to prevent QA data from referencing non-visualized data and to address other accuracy issues. This process involves feeding images, QA pairs, and reasoning chains into MLLMs to evaluate three key dimensions: chart relevance, data accuracy, and logical consistency.

\noindent\textbf{Difficulty Rating.} We perform 10 random sampling iterations using small MLLMs at high temperatures to establish clear difficulty levels based on chart complexity and task reasoning difficulty. The difficulty is quantified by the number of incorrect answers generated during these sampling runs, and overly simple questions are filtered out. For data intended for reinforcement learning, we further refine the selection to retain only "challenging but learnable" examples\cite{deepseekai2025deepseekr1incentivizingreasoningcapability}, ensuring optimal training effectiveness.

\noindent\textbf{Benchmark Refinement.} For the evaluation benchmark, we implement enhanced quality requirements beyond our standard pipeline. This included adjusting question difficulty distribution, conducting manual verification, and correcting. To ensure the benchmark effectively assesses models' genuine chart understanding capabilities, we use LLM as a judge to evaluate alignment between model predictions and answers. We also optimize judge prompts and eliminate questions that produce inconsistent evaluation results.

\begin{table}
  \centering
  \resizebox{\linewidth}{!}{%
  \begin{tabular}{lccc}
    \toprule
         \textbf{Statistic}    & \textbf{Train}  &  \textbf{Test} \\
    \midrule
    Total Questions             & 132,955 / 8,845      & 2,271 /  600    \\
    Chart Nums                   & 31,772 / 6,650      & 1,221 / 333    \\
    \midrule
     \textbf{Category}                              \\
     - Visual Recognition      & 56,651 / 0      & 681 /   0     \\
     - Data Extraction         & 23,680 / 2,963     & 501 / 200     \\
     - Calculation             & 21,614 / 2,861     & 593 / 200      \\
     - Data Analysis           & 19,609 / 3,021     & 496 / 200     \\
     - Chart2Markdown       & 11,401 / 0     & 0 / 0       \\
    \midrule
     \textbf{Tokens}                   \\
     - Avg Question      & 27.44 / 37.81     & 32.60 / 35.88     \\
     - Avg Reasoning     & 202.40 / 266.43   & 236.03 / 274.88    \\
     - Avg Answer        & 15.91 / 4.33    & 6.99 / 7.80      \\
    \bottomrule
  \end{tabular}
  }
  \caption{ChartM$^3$ dataset statistics with single-chart / multi-chart. The tokens of questions and answers are measured using Qwen2.5 tokenizer.}
  \label{tab:statistics}
\end{table}

Table~\ref{tab:statistics}  summarizes the statistics related to the final ChartM$^3$ dataset. Detailed quality control statistics and evaluation metrics are provided in Appendix~\ref{appendix:quality_assessment}.

\subsection{Chart RL with Verifiable Reward} \label{sec:35}
Studies involving DeepSeek-R1~\cite{deepseekai2025deepseekr1incentivizingreasoningcapability} and Kimi-1.5~\cite{kimiteam2025kimik15scalingreinforcement} have provided empirical evidence for the effectiveness of reinforcement learning with verifiable reward (RLVR) in improving the reasoning abilities of LLMs. Similarly, VLM-R1~\cite{shen2025vlmr1stablegeneralizabler1style} and R1-Omni~\cite{zhao2025r1omniexplainableomnimultimodalemotion} have extended this success to visual reasoning tasks. A key factor contributing to RLVR is the availability of large-scale data with verifiable answer formats, which enables effective reward modeling. Despite the promising results of RLVR in various domains, its application to chart understanding tasks remains unexplored mainly, with a notable scarcity of suitable datasets.

ChartM$^3$ offers an extensive collection of chart-text Q\&A pairs that naturally align with RLVR requirements. Leveraging this dataset, we propose a hybrid reward mechanism to adapt RLVR for chart understanding tasks. Following the Group Relative Policy Optimization (GRPO)~\cite{shao2024deepseekmathpushinglimitsmathematical} and reward modeling in DeepSeek-R1, our approach decomposes the reward signal into two components: accuracy reward $R_{acc}$ and format reward $R_{format}$, which are combined to form the total reward $R$.

The format reward $R_{format}$ evaluates whether the model's output adheres to the required output format: ``\texttt{<think>\{thinking process\}</think><answer>\{final answer\} </answer>}'', assigning a binary score (1 for compliance, 0 otherwise). The accuracy reward $R_{acc}$ incorporates both rule-based and model-based evaluation mechanisms:

\begin{itemize}
    \item \textbf{Rule-based reward:} For multiple-choice and true/false questions, we employ strict matching between the model predict and ground truth, yielding a binary reward (1 for exact match, 0 otherwise).
    
    \item \textbf{Model-based reward:} For fill-in-the-blank and short-answer questions, we use Qwen3-32B as a judge to evaluate response accuracy. The judge inputs the question, model's answer, and ground truth, producing a binary evaluation (1 for correct, 0 for incorrect).
\end{itemize}

\noindent Notably, CoT reasoning paths are not involved in the training process, with the model being optimized using only questions and final answers.

\section{Experiments}
\label{sec:experiments}

\begin{table*}[t]
\fontsize{14}{21}\selectfont
  \centering
  \resizebox{\textwidth}{!}{%
  \begin{tabular}{lccccccccccccccc}
\toprule[1.5pt]
\multirow{2}{*}{\textbf{Models}}  &  \multicolumn{6}{c}{\textbf{ChartM$^3$ test}} &  \multicolumn{4}{c}{ \textbf{ChartM$^3$-Multi test}} & \textbf{ChartQA*} & \textbf{ReachQA} & \multicolumn{1}{c}{\textbf{CharXiv}} \\
\cmidrule(lr){2-7} \cmidrule(lr){8-11} \cmidrule(lr){12-12} \cmidrule(lr){13-13}  \cmidrule(lr){14-14} 
&  \textbf{Overall} & \textbf{VR-A} &  \textbf{VR-B} & \textbf{ Ext.}  & \textbf{Calc.} & \textbf{ Ana.}  & \textbf{Overall} & \textbf{ Ext.} & \textbf{Calc.} & \textbf{Ana.} & \textbf{Overall} & \textbf{Overall} & \textbf{Overall}   \\

\midrule
\multicolumn{14}{c}{\textbf{Proprietary Multimodal Large Language Models}} \\
\midrule
Claude 3.5 Sonnet & \textbf{66.18} & \textbf{81.15} & \textbf{68.98} & \textbf{58.88} & \textbf{63.41} & \textbf{68.35} & \textbf{66.67} & \textbf{66.5} & \textbf{65.0} & \textbf{68.5} & \textbf{90.80} & \textbf{63.00} & \textbf{79.48} \\
GPT-4o & 58.30 & 78.53 & 63.67 & 48.90 & 53.12 & 60.89 & 53.33 & 50.0 & 46.5 & 63.5 & 86.70 & 53.25 & 76.98\\
GPT-4o mini & 48.35 & 82.20 & 54.08 & 39.52 & 39.97 & 48.59 & 42.50 & 38.0 & 39.0 & 50.5 & 77.52 & 40.35 & 66.76  \\

\midrule
\multicolumn{14}{c}{\textbf{Open-Source Multimodal Large Language Models}} \\
\midrule
Qwen2.5-VL-72B & \textbf{64.73} & \textbf{84.29} & \textbf{66.73} & \textbf{59.48} &  \textbf{60.37} & \textbf{65.73} & \textbf{61.00} & \textbf{59.0} & \textbf{59.0} & \textbf{65.0} & 88.60 & \textbf{61.55} & \textbf{82.24}  \\
InternVL3-78B & 55.57 & 77.49 & 62.24 & 51.30 & 46.88 & 55.24 & 45.50 & 44.0 & 40.5 & 52.0 & \textbf{89.60} & 47.25 & 80.00\\
Qwen2-VL-72B & 54.07 & 80.63 & 59.59 & 47.50  & 47.72 & 52.62 & 47.67 & 46.5 & 41.5 & 55.0 & 88.04 & 53.20 & 78.22\\

Qwen2.5-VL-7B & 57.42 & 79.06 & 59.18 & 50.10 & 52.78 & 60.28 & 52.00 & 48.5 & 46 & 61.5 & 87.60 & 57.65 & 67.50 \\
InternVL3-8B & 51.08 & 75.92 & 58.78 & 43.51 & 45.70 & 47.98 & 42.17 & 41.5 & 38.5 & 46.5 & 86.60 & 49.45 & 69.72 \\
InternVL2.5-8B & 42.10 & 66.49 & 51.02 & 36.93 & 29.01 & 44.76 & 36.50 & 29.0 & 29.5 & 51.0 & 77.60  & 35.20 & 63.20 \\
MiniCPM-V-2.6 & 40.64 & 68.59 & 46.94 & 32.14 & 30.02 & 44.96 & 34.67 & 32.0 & 26.5 & 45.5 & 79.20  & 34.65 & 51.86 \\

\midrule
\multicolumn{14}{c}{\textbf{ OCR/Chart-Augmented Open-Source Models}} \\
\midrule
mPlug-DocOwl2 & 23.25 & 32.98 & 15.71 & 20.76 & 13.83 & 40.73 & 23.17 & 16.0 & 13.0 & 40.5 & 66.64 & 10.90 & 26.74\\
ChartGemma & 22.99 & 45.55 & 15.71 & 22.75 & 14.5 & 31.85 & - & - & - & - & 71.28  & 18.50  & 18.84\\ 
TinyChart & 23.38 & 37.17 & 17.55 & 23.15 & 17.88 & 30.65 & 22.67 & 20.5 & 13.0 & 34.5 & 76.64 & 17.85 & 14.00\\

\midrule
\multicolumn{14}{c}{\textbf{ SFT Experiments on ChartM$^3$ with single and multi chart data}} \\
\midrule
Qwen2.5-VL-3B    & 45.00 & 65.45 & 45.31 & 44.51 & 36.59 & 47.38 & 34.83 & 32.0 & 25.0 & 47.5 & 83.92 & 45.75 & 54.34  \\
 + CoT-SFT         & \textbf{62.88} & \textbf{80.63} & \textbf{67.35} & \textbf{56.69} & \textbf{55.48} & \textbf{66.73} & \textbf{51.67} & \textbf{51.5} & \textbf{45.5} & \textbf{58.0} & \textbf{84.12} & \textbf{53.35} & \textbf{55.92} \\
\midrule
LLaVA-OV-7B & 37.12 & 63.35 & 42.86 & 29.34 & 24.96 & 43.75 & 29.00 & 27.0 & 17.5 & 42.5 & 80.44 & 28.40  & 46.24\\
 + CoT-SFT      & \textbf{64.95} & \textbf{83.25} & \textbf{68.98} & \textbf{63.47} & \textbf{57.50} & \textbf{64.31} & \textbf{54.33} & \textbf{53.5} & \textbf{50.0} & \textbf{59.5} & \textbf{82.32} & \textbf{43.40} & \textbf{51.04} \\

\bottomrule[1.5pt]
\end{tabular}%
}
  \caption{Evaluation results on ChartM$^3$ test set and other benchmarks. \textbf{Bold} values indicate the best performance within each category. Question categories names are abbreviated due to space limits. VR: Visual Recognition, Ext.: Data Extraction, Calc.: Calculation, Ana.: Data Analysis. "*" indicates that we use LLM as a judge to reevaluate ChartQA, which yielded slightly different results from those reported in the official technical report. Detailed explanations for LLM-based evaluation provided in the Appendix~\ref{appendix:judge_reason}.
  }
  \label{tab:benchmark}
\end{table*}

\subsection{Experimental Setup} \label{sec:setup}

\textbf{Baselines.} We evaluated three categories of MLLMs: (1) proprietary models, including GPT-4o \cite{jaech2024openai}, Claude3.5-Sonnet \cite{claude3.5-sonnet}, tested via official APIs. (2) Latest open-source models, including Qwen2-VL \cite{Qwen2-VL}, Qwen2.5-VL \cite{Qwen2.5-VL}, InternVL2.5 \cite{chen2024expanding},InternVL3 \cite{zhu2025internvl3exploringadvancedtraining}, LLaVA-OneVision \cite{li2024llava}, and MiniCPM \cite{yao2024minicpm}. (3) Open-source models specifically optimized for OCR and chart understanding, including mPlug-DocOwl2 \cite{hu2024mplugdocowl2highresolutioncompressingocrfree}, ChartGemma \cite{masry2024chartgemmavisualinstructiontuningchart}, TinyChart \cite{zhang2024tinychartefficientchartunderstanding}, and others. All models were evaluated using direct output (zero-shot inference)  with consistent default hyperparameters and prompts.

\noindent\textbf{Benchmarks.} Beyond ChartM$^3$ \texttt{test} set, we included established benchmarks for comparison: ChartQA \cite{masry-etal-2022-chartqa}, CharXiv \cite{wang2024charxiv}, ReachQA \cite{he2024distill}, SEED-Bench-2-Plus \cite{li2024seedbench2plusbenchmarkingmultimodallarge}, MMStar \cite{chen2024rightwayevaluatinglarge}, MathVista \cite{lu2024mathvistaevaluatingmathematicalreasoning}, and WeMath \cite{qiao2024wemathdoeslargemultimodal}. 
We adapted all benchmarks on VLMEvalKit \cite{duan2024vlmevalkit} and implemented accuracy evaluation using Qwen-Max \cite{qwen25} as the judge model, following their respective prompt designs.

\begin{table*}[t]
\fontsize{14}{21}\selectfont
  \centering
  \resizebox{\textwidth}{!}{%
  \begin{tabular}{lccccccccccccccc}
\toprule[1.5pt]
\multirow{2}{*}{\textbf{Models}}  &  \textbf{ChartM$^3$} & \textbf{ChartM$^3$-Multi} &  \multicolumn{3}{c}{\textbf{ChartQA*}} &  \multicolumn{3}{c}{\textbf{ReachQA}} &  \multicolumn{3}{c}{\textbf{CharXiv}} &  \multicolumn{4}{c}{\textbf{SEEDBench2\_Plus}} \\

 \cmidrule(lr){2-2} \cmidrule(lr){3-3}  \cmidrule(lr){4-6} \cmidrule(lr){7-9} \cmidrule(lr){10-12} \cmidrule(lr){13-16}
 & \textbf{Overall} & \textbf{Overall} &  \textbf{Overall} & \textbf{ Human} & \textbf{Aug.} & \textbf{Overall} & \textbf{ Reco.}  & \textbf{Reas.} & \textbf{ Overall} & \textbf{Desc.} & \textbf{ Reas.} & \textbf{Overall} & \textbf{Chart} & \textbf{Map} & \textbf{Web} \\

\midrule

Qwen2.5-VL-3B & 45.00 & 34.83 & 83.92 & 76.48 & 91.36 & 45.75 & 60.3 & 31.2 & 54.34 & 59.62 & 33.2 & 67.72 & 64.19 & 59.23 & 82.42 \\ 
 + CoT Prompt & 43.68 & 34.83 & 74.80 & 64.16 & 85.44 & 32.60 & 35.7 & 29.5 & 53.74 & 59.52 & 30.6 & 67.06 & 66.29 & 56.00 & 81.51 \\
 + SFT with 30K  data   & \textbf{58.17} & \textbf{47.17} & 82.20 & 75.84 & 88.56 & \textbf{50.10} & \textbf{60.8} & \textbf{39.4} & 54.44 & 60.6 & 29.8 & 66.13 & 64.81 & 55.14 & 81.21 \\
 + RL with 30K data     & 52.40 & 40.33 & \textbf{85.28} & \textbf{78.88} & \textbf{91.68} & 49.10 & 58.8 & \textbf{39.4} & \textbf{59.30} & \textbf{65.4} & \textbf{34.9} & \textbf{68.99} & \textbf{66.29} & \textbf{60.47} & \textbf{82.72} \\

\bottomrule[1.5pt]
\end{tabular}%
}
  \caption{ Reinforcement Learning results on five benchmarks. Details for these benchmarks are presented in \textsection~\ref{sec:setup}. \textbf{Bold} values indicate the best performance within each category.}
  \label{tab:rleval}
\end{table*}

\noindent\textbf{Training Evaluations.} To validate the effectiveness of ChartM$^3$, we first used Qwen2.5-VL as our base model and performed supervised fine-tuning (SFT) using our synthesized dataset of 142K training samples. We kept the vision encoder frozen while updating other modules, using a learning rate of 1e-5 and batch size of 64 for 2 epochs.

For RLVR experiment, the model was optimized with a learning rate of 1e-6 and KL divergence coefficient of 0.04. We sampled 7 rollouts for each prompt, and a global batch contained 7 different prompts. Considering both computational resource limitations and the importance of difficulty distribution in reinforcement learning training, we constructed our training set by sampling 30K items from the complete dataset according to their difficulty scores. More training and data selection details refer to the Appendix~\ref{appendix:grpo_setting}.

We utilized 8 NVIDIA A100 80G GPUs for all training process.

\subsection{Experimental Results}
\textbf{Our benchmark effectively measures chart comprehension and reasoning abilities.}  Both closed and open-source model evaluations show trends similar to ChartQA and ReachQA. Closed-source models demonstrate more balanced performance across all capability dimensions, while newer or larger open-source models exhibit stronger abilities across all test sets. Notably, ChartM$^3$-test significantly differentiates performance between various models. For instance, while models score above 86\% on ChartQA with minimal differences, ChartM$^3$-test reveals gaps exceeding 15\% between models like Claude 3.5 Sonnet (66.18\%) and InternVL3-8B (51.08\%).

\noindent\textbf{Existing advanced models excel at visual recognition but struggle with complex reasoning tasks.} Open-source models score significantly lower on complex reasoning tasks involving data extraction and computation compared to visual element recognition tasks, particularly evident in smaller-scale models. Additionally, we observed that some OCR/Chart-enhanced models perform well on ChartQA but struggle with ChartM$^3$-test and reasoning-intensive benchmarks. This disparity indicates their weakened instruction alignment and reasoning capabilities and suggests possible overfitting to traditional benchmarks.

\noindent\textbf{High-quality CoT data substantially improves chart reasoning performance.} As shown in Table~\ref{tab:benchmark}, CoT-SFT approach demonstrates substantial improvements, achieving at least 12\% performance gains over the base model on our benchmarks. The CoT-SFT model exhibits consistent improvements across both perception-oriented and comprehensive benchmarks in out-of-domain evaluations. Remarkably, on ReachQA, which demands complex reasoning capabilities, our CoT-SFT model achieves significant improvements of 7.60\% and 15.0\% over Qwen2.5-VL-3B and LLaVA-OV-7B, respectively. These substantial gains validate the quality of our dataset and its effectiveness in enhancing visual reasoning for universal chart understanding.

\noindent\textbf{Reinforcement Learning on ChartM$^3$ significantly improves both in-domain and out-of-domain performance.} As shown in Table~\ref{tab:rleval}, the model trained by GRPO obtains considerable improvement on various benchmarks. Compared to the base model, our RL approach yields notable gains in in-domain evaluations, achieving absolute improvements of 7.4\% and 5.5\% on ChartM$^3$ and ChartM$^3$-Multi benchmarks, respectively. In particular, the RL model demonstrates substantial improvements on out-of-domain benchmarks, particularly achieving a 4.96\% gain on CharXiv, suggesting better generalization capability than supervised fine-tuning. Further analysis on general and reasoning-specific benchmarks as shown in Table~\ref{tab:math_benchmarks} reveals that RL training preserves general capabilities (MMStar from 55.30\% to 56.00\%) while SFT shows potential decline. Notably, the RL model exhibits stronger performance on reasoning-intensive tasks, achieving a 5.14\% improvement on WeMath, suggesting effective transfer of learned reasoning patterns to broader analytical scenarios. This comprehensive improvement across diverse domains demonstrates the effectiveness of our synthetic datasets and training approach. 

\noindent\textbf{SFT and RL exhibit complementary strengths in chart understanding.} Our analysis reveals distinct advantages of SFT and RL approaches in different aspects of chart comprehension. SFT, driven by high-quality supervised signals, excels in perception-centric tasks by introducing new knowledge and extending vision-language alignment. In contrast, RL demonstrates superior capabilities in reasoning-intensive tasks by optimizing the probability of critical reasoning patterns, despite not introducing new knowledge. This complementary pattern is evidenced by their respective performance: while RL achieves moderate improvements in basic perception tasks, it shows substantial gains in complex reasoning scenarios by effectively discovering and strengthening crucial reasoning patterns.

These results validate that our synthetic chain-of-thoughts data successfully introduces diverse and essential patterns for complex chart understanding, effectively addressing scenarios where the base model lacks domain-specific knowledge.


\begin{table}
  \centering
  \resizebox{1\linewidth}{!}{%
    \begin{tabular}{l|ccc}
        \toprule[1.5pt]
        \textbf{Model} & \textbf{MMStar} & \textbf{MathVista} & \textbf{WeMath} \\
        \midrule
        Qwen2.5-VL-3B & 55.30 & 60.90 & 50.60 \\
        + SFT with 30K data & 53.70 & 55.30 & 51.20 \\
        + RL with 30K data & \textbf{56.00} & \textbf{61.60} & \textbf{55.74} \\
        \bottomrule[1.5pt]
    \end{tabular}
    }
    \caption{Performance comparison on general and math benchmarks.}
    \label{tab:math_benchmarks}
\end{table}

\subsection{Further Study}
In this subsection, we perform ablation studies to investigate the impact of different dataset compositions and training data sizes on the fine-tuning process.



\begin{table}
  \centering
  \resizebox{\linewidth}{!}{%
    \begin{tabular}{l|ccccc}
    \toprule[1.5pt]
    \textbf{Models} & \textbf{ChartM$^3$} & \textbf{ChartQA*} & \textbf{ReachQA} & \textbf{ChartXiv} \\
    \midrule
    Qwen2.5-VL-3B & 45.00  & 83.92 & 45.75 & 54.34 \\
    + ChartM$^3$ & \textbf{62.88}  & \textbf{84.12} & \textbf{53.35} & \textbf{55.92} \\
    + TinyChart & 42.18  & 81.60 & 42.60 & 51.40 \\
    + ChartGemma & 44.96  & 83.84 & 43.75 & 54.08 \\
    \bottomrule[1.5pt]
  \end{tabular}
  }
    \caption{Performance comparison of Qwen2.5VL-3B fine-tuned on different datasets.}
    \label{tab:dataset_comparison}
\end{table}

\noindent\textbf{ChartM$^3$'s Effectiveness over Existing Chart Datasets.}
To isolate the impact of dataset quality from model capability, we conducted controlled experiments using the same Qwen2.5-VL-3B baseline across ChartM$^3$ and existing datasets (ChartGemma and TinyChart), maintaining equal training samples and parameters. The results shown in Table ~\ref{tab:dataset_comparison} demonstrate that while ChartGemma showed minimal improvements and TinyChart even led to performance degradation, ChartM$^3$ achieved substantial gains across various benchmarks. 
This performance disparity underscores the significant challenge of enhancing chart comprehension capabilities on state-of-the-art models like Qwen2.5-VL, and validates that ChartM$^3$'s unique value stems from its comprehensive improvements in chart diversity, visual complexity, and high-quality Chain-of-Thought annotations, rather than from leveraging a more powerful base model.

\begin{figure}[t]
  \centering
  \includegraphics[width=0.96\linewidth]{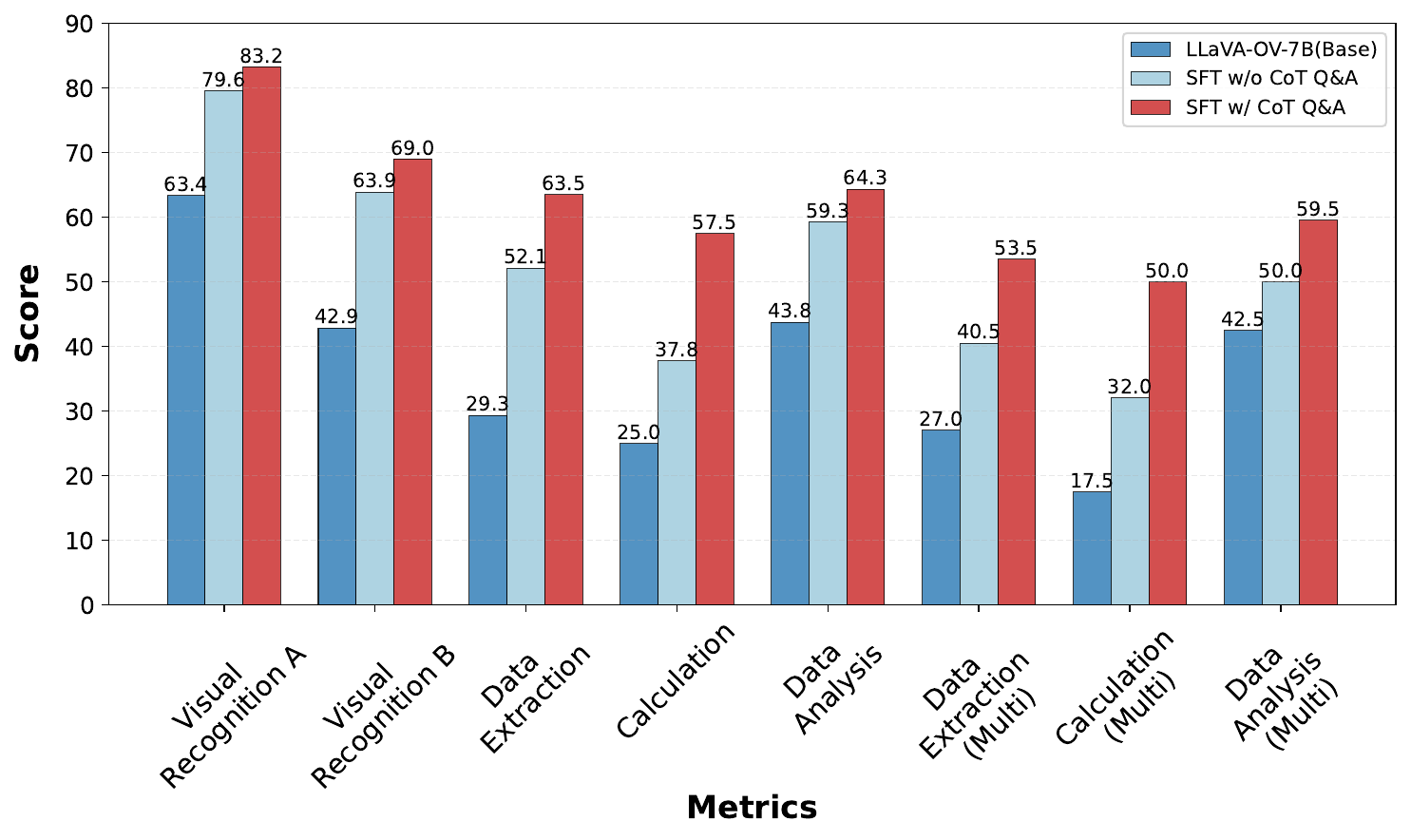}
  \caption{Performance comparison between models trained by SFT with and without CoT Q\&A across different evaluation metrics.}
  \label{fig:albation_dataset}
\end{figure}

\noindent\textbf{The Impact of CoT Data on Chart Reasoning Capabilities.}
Figure ~\ref{fig:albation_dataset} illustrates an ablation study on dataset composition by comparing models trained with and without CoT data. While both models achieve comparable performance on perception-based tasks, the CoT model significantly outperforms its counterpart on computation-intensive and statistic-related tasks, showing an 8\% performance improvement with the same amount of training data. These results demonstrate that high-quality CoT data serves as a key enabler for complex chart reasoning capabilities.

\begin{figure}[t]
  \centering
  \includegraphics[width=1\linewidth]{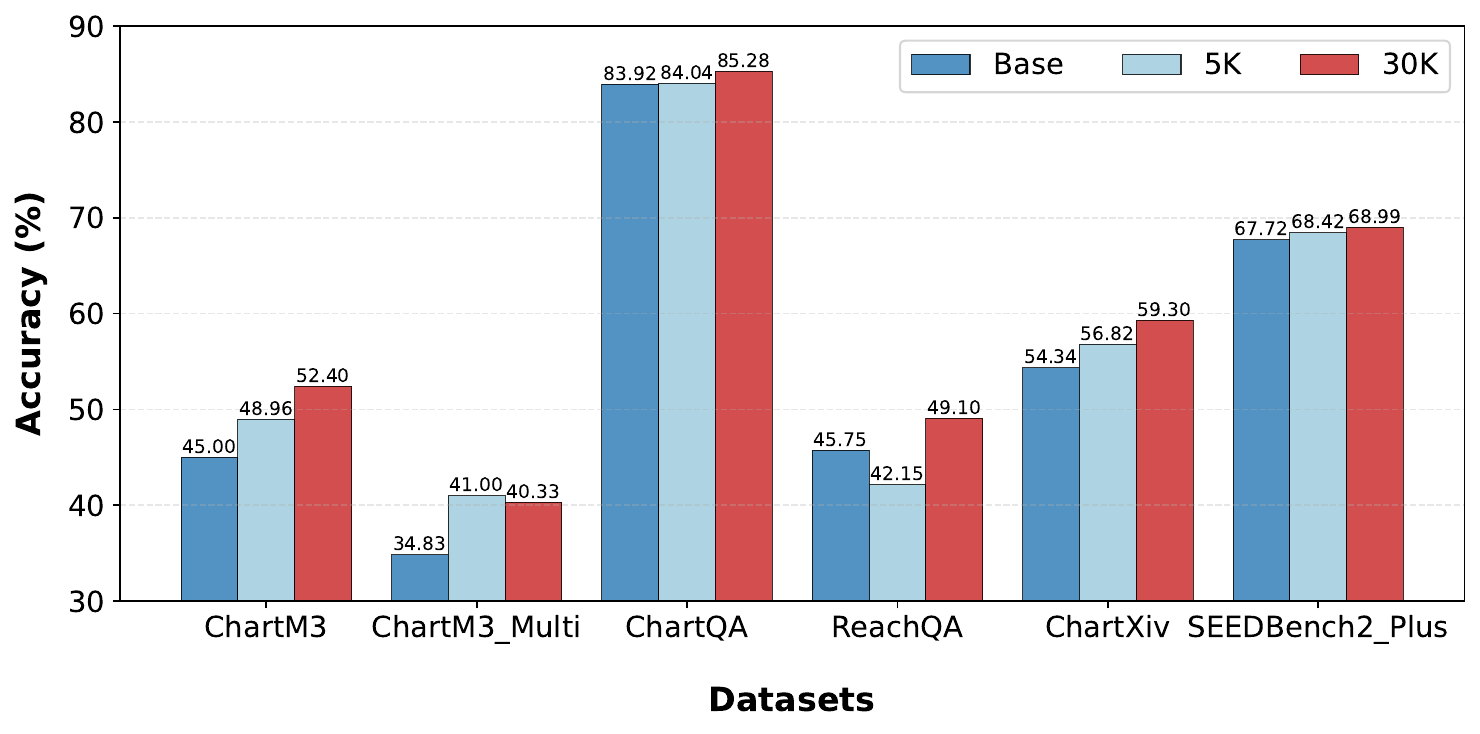}
  \caption{Performance of models trained by GRPO with different numbers of samples across multiple datasets.}
  \label{fig:data_scale}
\end{figure}

\noindent\textbf{The Impact of Training Data Scale on RL Performance.}
We conduct experiments with two different dataset sizes: 5,000 and 30,000 samples. As shown in Figure ~\ref{fig:data_scale}, the model trained with 30,000 samples consistently outperforms its counterpart trained with 5,000 samples across most datasets. While reinforcement learning is generally considered data-efficient, scaling up training data leads to substantial improvements. This is particularly crucial for fill-in-the-blank and short-answer questions, where beneficial reasoning patterns are more sparse and require larger datasets to be effectively captured during training. Notably, with limited training data (5K samples), the model's performance on ReachQA degrades due to the high variance nature of RL training, but this instability is effectively addressed when scaling up to 30K samples, yielding a 6.95\% improvement.

\section{Conclusion}
\label{sec:conclusion}

This work examines current MLLMs' challenges in real-world chart comprehension and evaluates the limitations of existing dataset construction methods. We propose a multi-stage, code-driven pipeline for synthesizing visual reasoning Q\&A data. Our method starts by generating a key question, retrieving appropriate chart templates, using LLMs to generate code that simulates real data distribution, plotting charts and solving problems, and implementing data filtering through various-sized models to obtain diverse charts and high-quality CoT data. We have developed ChartM$^3$, a multi-dimensional and multi-step dataset, and conduct CoT supervised fine-tuning and reinforcement learning. The results show significant performance improvements across multiple benchmarks. Our framework bridges the gap between academic research in chart understanding and practical applications, advancing the development of reasoning MLLMs.

\section*{Limitations}

Although our work achieves promising results in chart-related reasoning tasks, several limitations exist. (1) The chart rendering code is primarily Python-based, with limited support for other visualization languages, suggesting a need to incorporate additional languages to diversify chart generation capabilities. (2) This work concentrates mainly on statistical charts. Future research should consider extending this approach to flowcharts (such as process diagrams and relationship diagrams) and other visual formats. (3) The reinforcement learning experiments are not conducted at a larger scale. In the future, we will explore expanding the data scale, model size, and investigating chart reasoning data distillation based on reinforcement learning.

\section*{Ethical Consideration}
 
We strictly declare that all authors are aware of and adhere to the ACL Code of Ethics throughout this research. We strictly adhere to the licenses of all open source datasets and models used. During the benchmark refinement phase of Data Evaluation, quality validation was conducted through human annotations. Annotators received task-specific materials and explicit consent was obtained for using their annotations exclusively for academic research purposes. It is imperative to ensure the privacy of all annotators throughout the annotation process. Furthermore, all annotators were adequately compensated according to local standards.

For this work, we used open-source and closed-source models obtained from official sources and accessible to the public to avoid potential harm to individuals or groups. We did not use any personally identifiable information, and all data were anonymized before analysis. The prompts and benchmarks underwent a meticulous human selection and processing phase to ensure no names or unique identifiers of individual people or offensive content were included. Additionally, we used Grammarly to refine the language in our manuscript.

\bibliography{ref.bib}

\appendix

\section{Appendix}
\label{sec:appendix}

\subsection{Data Categories}

\begin{figure}[t]
  \centering
  \includegraphics[trim=5mm 30mm 5mm 30mm, clip, width=1\linewidth]{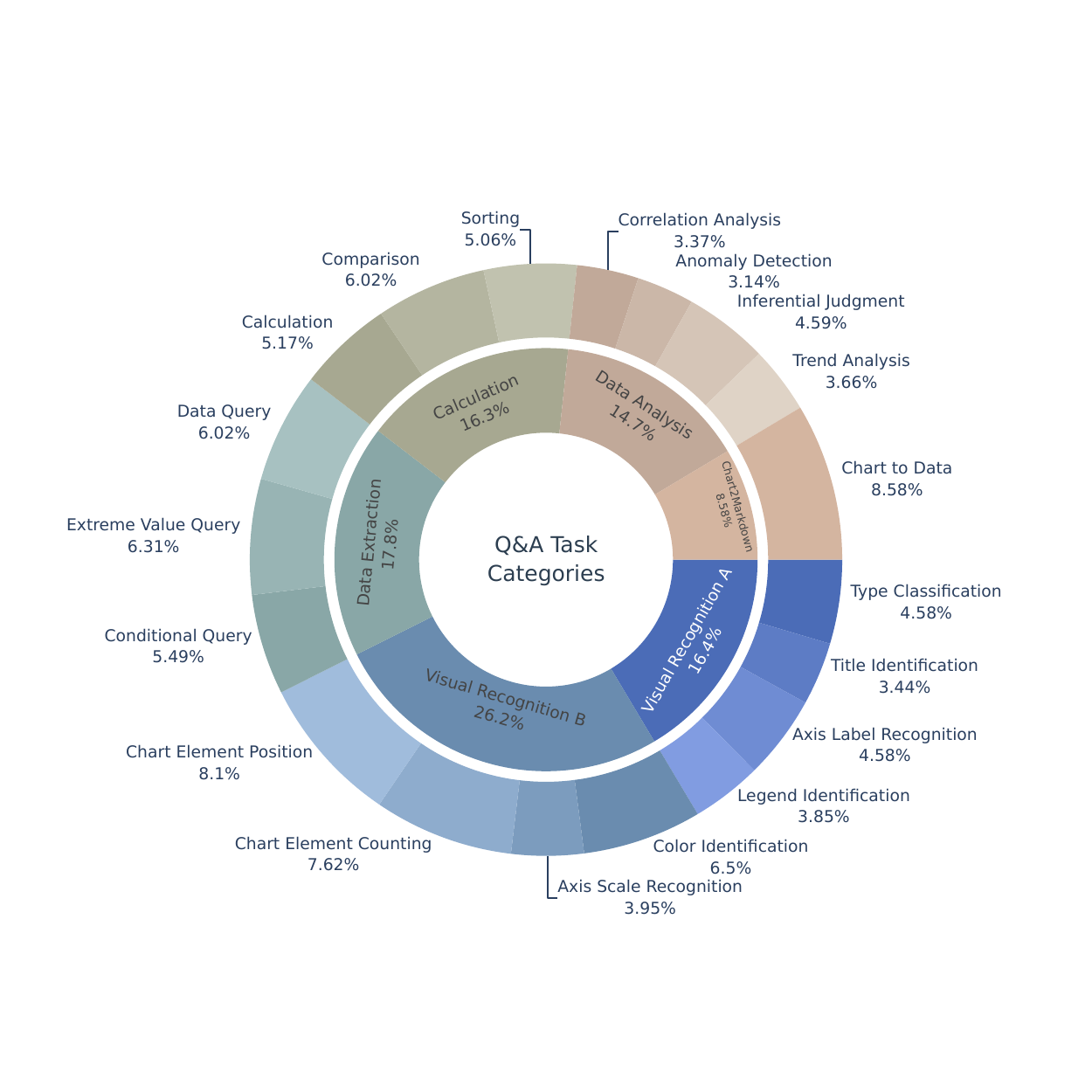}
  \caption{The distribution of ChartM$^3$ Q\&A categories.}
  \label{fig:subcategories}
\end{figure}

In our generation pipeline, we predefine chart types, Q\&A task categories, and visualization domains. Table~\ref{tab:chart_types} presents 9 major, 62 minor chart types. Table~\ref{tab:task_categories} outlines 18 specialized Q\&A categories across 4 primary dimensions, along with the Chart To Markdown task. Due to varying difficulty levels, we have divided Visual Recognition into two parts: A and B. The distribution of questions across these subcategories is illustrated in Figure~\ref{fig:subcategories}. Additionally, Table~\ref{tab:chart_domains} enumerates 60 domains commonly used in data visualization.

\subsection{Dataset Quality Assessment}
\label{appendix:quality_assessment}

We conducted comprehensive quality control processes for both chart images and Q\&A pairs. Table~\ref{tab:quality_control} presents the filtering statistics across different components of our dataset.

For chart quality verification, we developed a classifier using Qwen2-VL-2B trained on manually curated examples. Table~\ref{tab:quality_metrics} shows the classifier's performance on a validation set of 107 instances.

To assess instruction accuracy, we evaluated approximately 5,800 samples using Claude 3.5, followed by dual-verification (combining Claude 3.5 and human expertise) for cases with incorrect responses. This process identified 508 instances requiring modification or removal, resulting in an instruction accuracy of 91.24\%.

\begin{table}[t]
\centering
\resizebox{1\linewidth}{!}{
\begin{tabular}{lccc}
\toprule
\textbf{Dataset} & \textbf{Initial} & \textbf{Reserved} & \textbf{Rate(\%)} \\
\midrule
\multicolumn{4}{l}{\textbf{ChartM$^3$}} \\
Chart Quality & 38,452 & 34,064 & 88.59 \\
Q\&A Quality & 171,531 & 140,312 & 81.80 \\
\midrule
\multicolumn{4}{l}{\textbf{ChartM$^3$-Multi}} \\
Chart Quality & 4,336×2 & 3,777×2 &  87.11 \\
Q\&A Quality & 11,331 & 9,821 & 86.68 \\
\bottomrule
\end{tabular}
}
\caption{Statistics of quality control filtering process. Note that each data point in ChartM$^3$-Multi contains two charts.}
\label{tab:quality_control}
\end{table}

\begin{table}[t]
\centering
\resizebox{1\linewidth}{!}{
\begin{tabular}{lccc}
\toprule
\textbf{Category} & \textbf{Precision(\%)} & \textbf{Recall(\%)} & \textbf{F1-score(\%)} \\
\midrule
Low Quality & 93.33 & 87.50 & 90.32 \\
High Quality & 90.32 & 94.92 & 92.56 \\
\bottomrule
\end{tabular}
}
\caption{Performance metrics of the chart quality classifier.}
\label{tab:quality_metrics}
\end{table}

\begin{table}[t]
\small
\centering
\begin{tabular}{lc}
\toprule
\textbf{Question Type} & \textbf{Count} \\
\midrule
True/False & 6,958 \\
Multiple-choice & 6,734 \\
Short-answer & 2,657 \\
Fill-in-the-blank & 13,651 \\
\bottomrule
\end{tabular}
\caption{Distribution of different question types in GRPO training dataset.}
\label{tab:question_types}
\end{table}

\subsection{GRPO Training Setting}
\label{appendix:grpo_setting}
\textbf{Data Sampling for GRPO.} DAPO~\cite{yu2025dapoopensourcellmreinforcement} indicates that samples with zero advantage variance lead to performance degradation, thus should be filtered out during training. Based on this finding, we carefully curate the GRPO training dataset by filtering out both overly difficult and simple samples. Specifically, we perform uniform sampling from items with difficulty scores ranging from 3 to 9 (difficulty score definition in Section~\ref{sec:34}) to ensure a balanced distribution of task complexity. Additionally, we maintain an approximately 1:1 ratio between questions with rule-based rewards (True/False and Multiple-choice) and model-based rewards (Short-answer and Fill-in-the-blank), as shown in Table~\ref{tab:question_types}.

\noindent\textbf{KL Divergence Approximation.} In original GRPO, KL divergence approximation can be formulated as Eq.~\ref{eq:k3}:
\begin{equation}
\begin{aligned}
\mathbb{D}_{KL}[\pi_\theta\Vert\pi_{ref}] &= r - \log r - 1, \\
\text{where } r &= \frac{\pi_{ref}(a|s)}{\pi_\theta(a|s)}
\end{aligned}
\label{eq:k3}
\end{equation}
where $a$ denotes the current token and $s$ represents previous sequence before $a$, $\pi_{ref}$ is the reference model initialized from base model, and $\pi_{\theta}$ is the policy model being optimized.

In this paper, all GRPO experiments apply another approximation, called $k2$~\cite{schulman2021kl}, and can be formulated as Eq.~\ref{eq:k2}:
\begin{equation}
\mathbb{D}_{k2}[\pi_\theta\Vert\pi_{ref}] = \frac{1}{2}(\log r)^2
\label{eq:k2}
\end{equation}
where $r$ is defined the same as in Eq.~\ref{eq:k3}.

\subsection{Explanation for LLM-based Evaluation}
\label{appendix:judge_reason}
This work utilizes LLM-based evaluation for all chart benchmarks, including ChartQA. 
The traditional evaluation method for ChartQA, which relies on string exact matching and numerical calculations within a relative error range, exhibits  several limitations:

\begin{enumerate}
    \item Unit Discrepancies: Mismatches occur when predicted results include units while reference answers do not (for example, "5" versus "5 meters" or "5" versus "5 million").
    \item Numerical Range Issues: When labels on the x-axis are numbers (particularly years), the traditional evaluation method's 5\% error range is too permissive. For instance, if the correct answer is 2000, predictions ranging from 1900 to 2100 would be incorrectly marked as correct.
\end{enumerate}

These limitations make it difficult to accurately assess the performance of MLLMs that have not been specifically trained on similar data distributions. To address these issues, our experiment employs LLMs as judges, resulting in more accurate evaluations. The detailed judge prompt is shown in Figure~\ref{fig:judge_prompt}. 

Meanwhile, to ensure more comprehensive evaluation and alignment with previous works, we expanded our evaluation framework to include the original Relaxed Accuracy metric as used in previous works, an enhanced version of Relaxed Accuracy (which automatically removes units for numerical answers and standardizes number formatting, such as converting "116,000" to "116000") for ChartQA, and GPT-4o (gpt-4o-2024-11-20) as a judge for CharXiv. Performance comparison among different evaluation metrics is shown in Table~\ref{tab:model_comparison}. 

\begin{table*}[t]
\centering
\resizebox{0.9\linewidth}{!}{
  \begin{tabular}{l|ccc|cc}
\toprule[1.5pt]
\multirow{2}{*}{\textbf{Models}}  & \multicolumn{3}{c|}{\textbf{ChartQA}} & \multicolumn{2}{c}{\textbf{CharXiv}} \\
& \makecell{\textbf{Oral} \\ \textbf{Relaxed Acc.}} & \makecell{\textbf{Advanced} \\ \textbf{Relaxed Acc.}} & \textbf{QwenMax} & \textbf{GPT-4} & \textbf{QwenMax} \\
\midrule

Qwen2.5-VL-3B & 83.16 & 83.64 & 83.92 & 53.14 & 54.34 \\
+ CoT-SFT with 142K data & 78.16 & 84.56 & 84.12 & 54.02 & 55.92 \\
LLaVA-OV-7B & 80.72 & 81.08 & 80.44 & 45.10 & 46.24 \\
+ CoT-SFT with 142K data & 72.04 & 82.00 & 82.32 & 49.18 & 51.04 \\
\midrule
Qwen2.5-VL-3B & 83.16 & 83.64 & 83.92 & 53.14 & 54.34 \\
+ CoT-SFT with 30K data & 79.64 & 82.76 & 82.20 & 52.74 & 54.44 \\
+ RL with 30K data & 79.52 & 85.32 & 85.28 & 57.82 & 59.30 \\

\bottomrule[1.5pt]
\end{tabular}%
}

  \caption{Performance comparison across different models and training approaches on ChartQA and CharXiv datasets using various evaluation metrics. Acc.: Accuracy.}
  \label{tab:model_comparison}
\end{table*}

\subsection{Examples of Chart Template Database}

\begin{figure*}[t]
  \centering
  \includegraphics[width=1\linewidth]{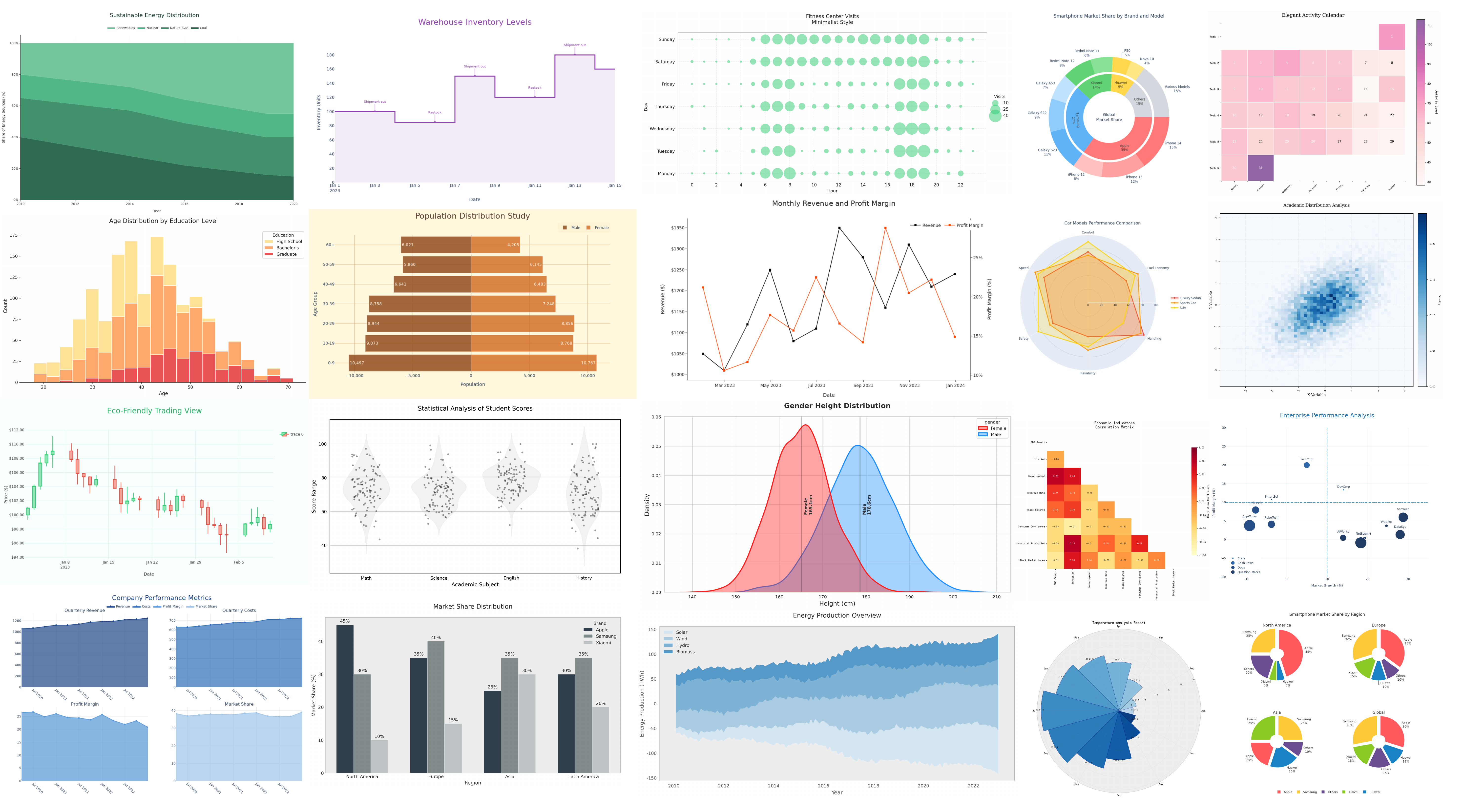}
  \caption{Examples of ChartM$^3$ Template Database.}
  \label{fig:template_visualization}
\end{figure*}

We sample several charts from ChartM$^3$ chart template database. The visualization is presented in Figure~\ref{fig:template_visualization}.

\subsection{Examples of Evaluation Comparisons}
\begin{figure*}[t]
  \centering
  \includegraphics[width=1\linewidth]{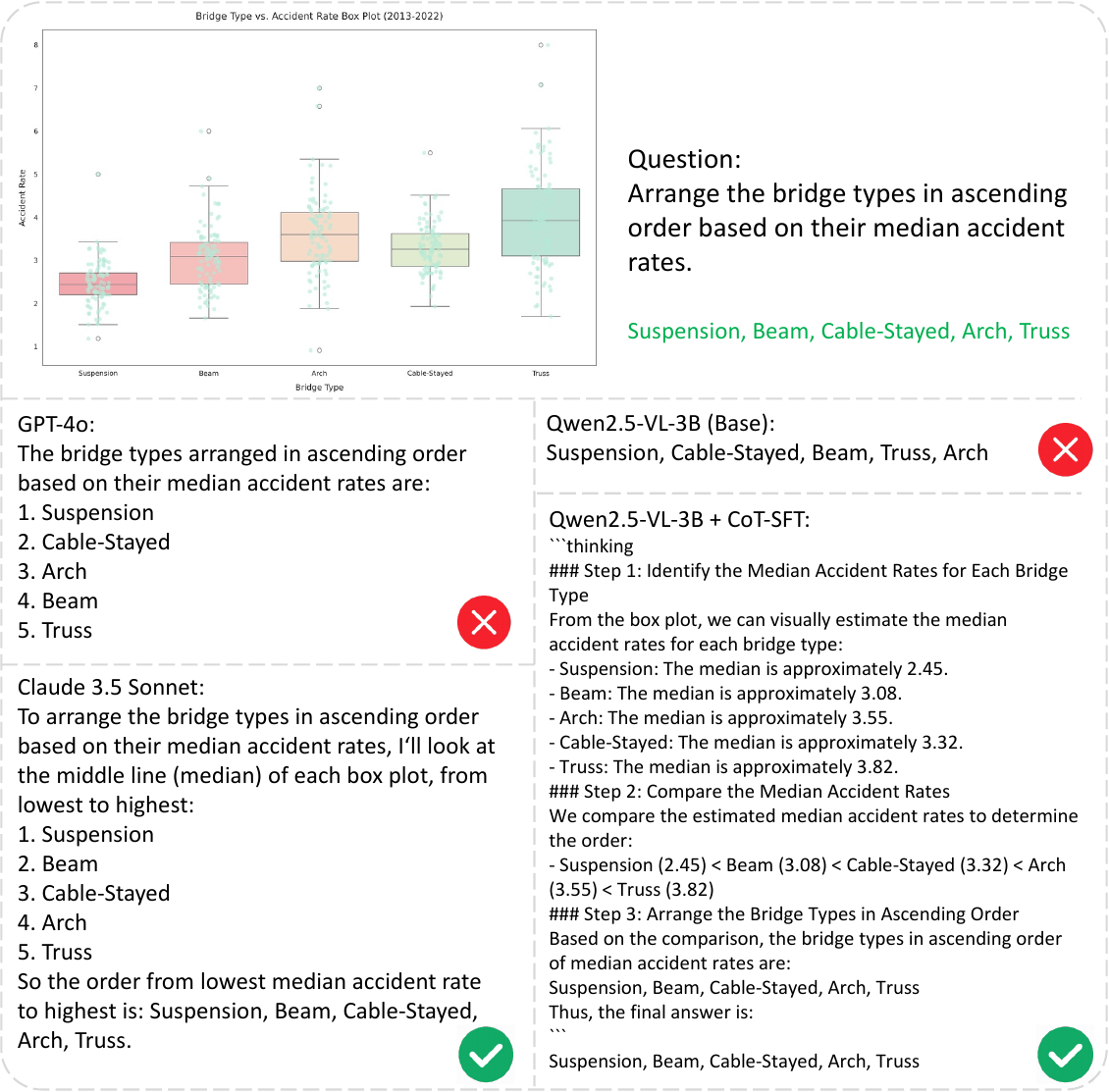}
  \caption{A Case Study of ChartM$^3$ Evaluation Results. While both GPT-4o and the base model provided incorrect answers, the model trained with CoT-SFT successfully analyze the medians across categories during its reasoning process and produce the correct ranking.}
  \label{fig:demo1}
\end{figure*}

\begin{figure*}[t]
  \centering
  \includegraphics[width=1\linewidth]{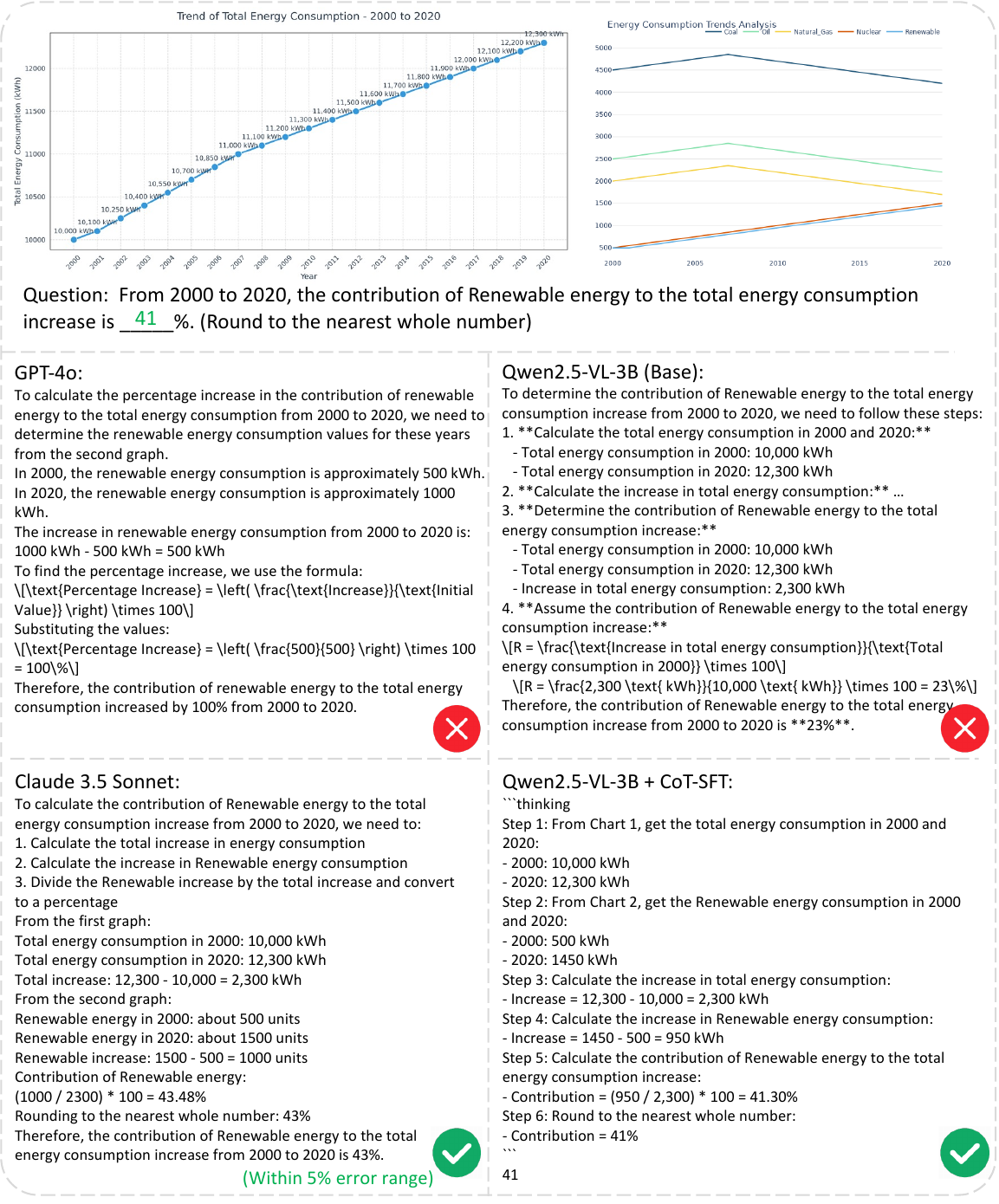}
  \caption{A Case Study of ChartM$^3$ Evaluation Results for Multi-Chart Scenarios. Although individual chart elements are straightforward, GPT demonstrates limitations in cross-graph analysis. Specifically, when examining renewable energy growth from 2000 to 2020, GPT fails to properly reference the first graph. The base model incorrectly substitutes total energy consumption data for renewable energy consumption. In comparison, the model trained with CoT-SFT correctly identifies that renewable energy levels in 2020 are below 1500 units, producing a prediction that more closely aligns with the standard answer compared to Claude 3.5 Sonnet.}
  \label{fig:demo3}
\end{figure*}

We provide comparative examples of multiple models' evaluation results on ChartM$^3$ to demonstrate that after Chain-of-Thought Self-Fine-Tuning (CoT-SFT) with high-quality data, the base model significantly improves reasoning capabilities in complex chart comprehension. The examples of the evaluation results are presented in Figure~\ref{fig:demo1} and Figure~\ref{fig:demo3}.

\subsection{Prompt Templates}
\label{appendix:prompt_templates}
We present the prompt templates used in this paper.

\textbf{Prompt for Data Generation.} We utilize LLMs to transform the key questions into realistic contextual narratives and output data generation code rather than direct data. The prompt is shown in Figure~\ref{fig:data_prompt}.

\textbf{Prompt for Visualization Generation.}  We employ a step-by-step reasoning approach to improve code usability and visual presentation. The process begins by guiding LLMs through visualization requirement analysis and developing a detailed solution of visual elements. Using the solutions as few-shot prompt, we generate and execute visualization code to create chart images. The prompts are shown in Figure~\ref{fig:vis_prompt_1} and Figure~\ref{fig:vis_prompt_2}.

\textbf{Prompt for Q\&A Generation.} We employ a two-stage Code-driven approach for Q\&A pair construction. The first stage involves question formulation and analytical code synthesis for each question and its source data. The second stage generates CoT reasoning and precise answers through code execution results and the computational process. The prompts are shown in Figure~\ref{fig:qa_prompt_1} and Figure~\ref{fig:qa_prompt_2}.

\textbf{Prompt for Evaluating Models.} In the evaluation of ChartM$^3$, we use Qwen-Max as the judge model, the judge prompt is optimized based on Reachqa and CharXiv methods, which is shown in Figure~\ref{fig:judge_prompt}. 
\label{appendix:judge_Prompt}

\begin{table*}
    \centering
  \resizebox{\textwidth}{!}{%
  \begin{tabular}{ll}
    \hline
    \textbf{Major Category} & \textbf{Minor Category}  \\
    \hline
    Bar & Single Bar Chart, Grouped Bar Chart, Stacked Bar Chart, Positive-Negative Bar Chart, \\
    & Lollipop Plot, Bidirectional Bar Chart, Butterfly Diagram, Range Bar Chart,  \\
    & Waterfall Plot, Candlestick Plot, Single Histograms, Rectangular Funnel Chart, Box Plot,  \\
    & Error Bars Chart, Bullet Chart, Barbell Chart, Nested Bar Chart, Dumbbell Plot   \\

    \midrule
    
    Line & Single Line Chart, Grouped Line Chart, Stacked Line Chart, Slope Graph, Step Chart \\

    \midrule
    
    Area & Single Area Chart, Stacked Area Chart, Bilateral Area Chart, Range Area Chart, Streamgraph,  \\
    & Error Bands Chart,  Density Plot \\

    \midrule
    
    Pie & Single Pie Chart, Multidimensional Pie Chart, Donut Pie Chart, Multilevel Donut Chart, \\
    & Sunburst Chart \\

    \midrule
    
    Radar & Single Radar Chart, Grouped Radar Chart, Stacked Radar Chart, Single Rose Chart,  \\
    &  Grouped Rose Chart, Stacked Rose Chart \\

    \midrule
    
    Scatter & Scatter Plot, Bubble Plot, Quadrant Plot, Strip Plot, Swarm Plot, Violin Plot \\

    \midrule
    
    Heatmap & Heatmap Plot, Calendar Heatmap, Waffle Chart \\

    \midrule
    
    Progress & Gauge graph, Semi-circular Progress Chart, Bar Progress Chart, Circular Progress Chart \\

    \midrule
    
    Combination & Line-Column Combination Chart, Line-Area Combination Chart, Dual Y-Axis Line Chart, \\
    & Dual Y-Axis Bar Chart, Multiple Subplot Bar Chart, Multiple Subplot Area Chart, \\
    & Multiple Subplot Line Chart, Multiple Subplot Pie Chart \\

    \hline
  \end{tabular}}
  \caption{ Major and Minor Charts Types.
  }
  \label{tab:chart_types}
\end{table*}

\begin{table*}
  \centering
  \resizebox{\textwidth}{!}{%
  \begin{tabular}{ll}
    \hline
    \textbf{Major Category} & \textbf{Minor Category}  \\
    \hline
    Visual Recognition A & Type Classification,  Title Identification, Axis Label Recognition, Legend Identification\\
    Visual Recognition B & Color Identification,  Axis Scale Recognition, Chart Element Counting, Chart Element Position\\
    Data Extraction & Data Query,  Extreme Value Query, Conditional Query\\
    Calculation & Calculation, Comparison,  Sorting\\
    Data Analysis & Correlation Analysis, Anomaly Detection, Inferential Judgment, Trend Analysis \\
    Chart2Markdown & Chart To Markdown\\
    
    \hline
  \end{tabular}}
  \caption{ Major and Minor Categories of Charts. }
  \label{tab:task_categories}
\end{table*}

\begin{table*}
  \centering
  \resizebox{\textwidth}{!}{%
  \begin{tabular}{lllll}
    \hline
    Education & Art & Finance & Healthcare & Information Technology \\
    Environmental Science & Social Science & Economics & Political Science & History \\
    Psychology & Management & Marketing & Law & Engineering \\
    Physics & Chemistry & Biology & Geography & Astronomy \\
    Geology & Meteorology & Oceanography & Agriculture & Forestry \\
    Animal Husbandry & Fishery & Food Science & Energy & Materials Science \\
    Mechanical Engineering & Electrical Engineering & Civil Engineering & Aerospace & Transportation \\
    Architecture & Urban Planning & Interior Design & Industrial Design & Fashion Design \\
    Graphic Design & Advertising & Journalism & Public Relations & Sports Science \\
    Entertainment & Tourism  & Retail & Manufacturing & Logistics \\
    Human Resources & Corporate Strategy & Risk Management & Audit \& Accounting & Tax \\
    Non-profit Management & International Relations & Foreign Policy & Hospitality & Supply Chain\\

    \hline
  \end{tabular}}
  \caption{Chart Domains. }
  \label{tab:chart_domains} 
\end{table*}

\begin{figure*}
\begin{promptblock}[LLM Prompt]{lightblue}{prompt:datagen}

You are a senior business analyst and data visualization expert. Please generate high-quality data for chart creation based on the following detailed requirements.
The generated data should solve a key question through chart visualization. You need to first conceive a realistic background story based on the specified chart type, business domain, theme, and other conditions, then provide the data generation code.

\vspace{\baselineskip}
\#\# Basic Information Requirements

1. Key Question: \{key\_question\}

2. Domain: \{domain\}

\vspace{\baselineskip}
\#\# Chart Type Information 

Here is the specific information of chart type: \{description\}

\vspace{\baselineskip}
\#\# Data Content Requirements

1. Data Description: 

   - Data background overview (time range, data source, etc.)
   
   - Data distribution and overall trend analysis
   
   - Key feature points explanation (maximum, minimum, turning points, etc.)
   
   - Comparative analysis between data
   
2. Chart Title

   - Title should be concise and summarize core information
   
   - Include key dimensional information (time, location, object, etc.)
   
   - For stacked charts, specify chart type in the title

3. Original Data Generation Code

   - Python code, import necessary libraries like import pandas as pd and import numpy as np
   
   - Can use random numbers and mathematical distribution functions to generate data
   
   - Save all data as data.csv file, first row must be column names
   
   - Ensure generated values retain maximum three significant digits
      
   - Ensure code is executable correctly
   
\vspace{\baselineskip}
\#\# Data Generation Rules 

1. Data Structure Requirements:

   - Ensure data structure fully complies with technical requirements of specified chart type
   
   - Data scale should be reasonably set while maintaining chart clarity and readability
   
   - All data items must contain complete label information

2. Data Quality Requirements:

   - Choose appropriate data distribution and trends based on actual business domain characteristics
   
   - Unless specifically required in key question, legends should not exceed 5
   
   - Value ranges must be reasonable and business meaningful
   
   - If including time series, ensure consistency of time intervals
   
   - Can include 1-2 meaningful outliers, but proportion should not exceed 10\% of total data

3. Business Background Requirements:

   - Provide detailed data collection background (time range, geographic range, statistical criteria, etc.)
   
   - Fictional details need to maintain internal consistency
   
   - All value changes should be explainable by business logic
   
\vspace{\baselineskip}
\#\# Common Data Distribution References 

   Normal distribution,
   Poisson distribution,
   Uniform distribution,
   Exponential distribution,
   Skewed distribution,
   Multi-modal distribution,
   Long-tail distribution,
   Bimodal distribution,
   Other distributions,

\vspace{\baselineskip}
\#\# Common Data Trend References 

   Linear trends(continuous rise, continuous fall, stable),
   Cyclical trends,
   Compound trends,
   Mutation patterns,
   Fluctuation patterns,
   S-curve,
   Other trends,

\vspace{\baselineskip}
\#\# Data Generation Code Example 

\{example\_data\}

\vspace{\baselineskip}
\#\# Output Format 

Output all content in English.

First provide the thinking process, output in a code block with "thinking" header. Then output the result in JSON format without any other content, including the following fields:

\{
"description": "Data description",
"title": "Chart title",
"data\_code": "Original data generation code"  
\}

\end{promptblock}
    \caption{Prompt template for data generation.}
  \label{fig:data_prompt}
\end{figure*}

\begin{figure*}
\begin{promptblock}[LLM Prompt]{lightblue}{prompt:vis_prompt_1}

You are a data visualization expert responsible for analyzing visualization requirements and providing detailed chart design recommendations. Please analyze according to the following steps based on user requirements and uploaded data.

\vspace{\baselineskip}
Phase 1: Requirements Analysis, consider the following questions:

1. Data Analysis

- What are the key characteristics of the provided data?

- Which relationships or patterns need to be highlighted?

2. Background Understanding  

- What is the industry background and target audience?

- What insights need to be conveyed?

- What are common visualization methods in this field?

3. Visualization Strategy, based on data characteristics and business context:

- Which chart types are most effective?

- What alternatives were considered and why were they rejected?

- If needed, how should multiple elements be composed?

\vspace{\baselineskip}
Phase 2: Visualization Design, develop visualization solutions based on above results.

1. Detailed Design Specifications for implementation in Python visualization libraries like Matplotlib or plotly. Pay attention to chart aesthetics:

- Chart type and layout [User selected chart type: \{target\_chart\_type\}, do not consider other types]

- Color scheme and style

- Axis configuration and scale

- Labels, titles and annotations [Note: All text content (titles, legends, axis labels etc.) should be in English]

- Legend position and format  

- Gridlines and other reference elements

- Size and aspect ratio

- Other visual elements

Note: All above content must be designed only when relevant data columns exist. Do not generate plotting requirements without data conditions!

\vspace{\baselineskip}
Below are the user data characteristics and requirements:

\vspace{\baselineskip}
\#\# User Data Start 

Title: \{file\_name\}

Goal: \{seed\_description\}

data.head():
\{data\_head\}

data.describe():
\{data\_describe\}

data.describe(include='object'):
\{data\_describe\_object\}

\#\# User Data End

\vspace{\baselineskip}
Now, please begin analysis and output a JSON string in a ```json code block containing these two fields (both plain text, add line breaks between points):

- 'analysis': Provide thought process for requirements analysis phase

- 'guidance': Provide visualization design phase solutions (note: no actual visualization code needed)
Do not output anything besides JSON. Keep results concise and refined without excessive verbiage.

\end{promptblock}
    \caption{Prompt template for the first stage in visualization generation.}
  \label{fig:vis_prompt_1}
\end{figure*}

\begin{figure*}
\begin{promptblock}[LLM Prompt]{lightblue}{prompt:vis_prompt_2}

You are a data visualization expert with a Python visualization code generation task. You need to first read the example code, then implement visualization code for user data based on their requirements.

\vspace{\baselineskip}
\#\# Example Start

Target Chart Type: \{target\_chart\_type\}
\{visual\_definition\}

Sample Data Format:
\{sample\_data\_head\}

Sample Plot Code:
\{sample\_code\}

\#\# Example End

\vspace{\baselineskip}
Below are the user data characteristics and requirements:

\#\# User Data Start

Title: \{file\_name\}

Goal: \{seed\_description\}

data.head():
\{data\_head\}

data.describe():
\{data\_describe\}

data.describe(include='object'): 
\{data\_describe\_object\}

\#\# User Data End

\vspace{\baselineskip}
Actual Visualization Requirements:
\{vis\_guidance\}

All text content in charts (titles, legends, axis labels etc.) should be in English.

Now, please reference the example and generate visualization code meeting the requirements based on actual user data situation and needs.

\vspace{\baselineskip}
Specific requirements:

1. User data is loaded into memory in 'data' variable as pandas.DataFrame. Do not output any data reading/declaration code.

2. Based on example code, try to meet actual visualization requirements but avoid complex code modifications to prevent errors. For long text, avoid overlapping text in x-axis, legend etc.

3. Generate two Python functions: 'def preprocess(data):' for plot data preprocessing, input is raw dataframe, output is preprocessed dataframe; 'def plot(data):' for drawing corresponding charts. Only generate one final chart (can have multiple subplots).

4. preprocess function needs to be called in plot function. Only generate function bodies, no need for plot function calling code.

5. Complete all plot data preprocessing in preprocess function (including decimal places), no data processing in plot function!

6. Save result to file named 'plot.png'.

7. Most importantly, ensure code can execute correctly, so keep plotting function parameters consistent with example as much as possible. Generate all code in one ```python code block.

\end{promptblock}
    \caption{Prompt template for the second stage in visualization generation.}
  \label{fig:vis_prompt_2}
\end{figure*}

\begin{figure*}
\begin{promptblock}[LLM Prompt]{lightblue}{prompt:qagen1}
You are a senior business analyst with extensive experience in data analysis and visualization. Your task is to generate a high-quality analytical question based on chart visualization code and data, and write Python code to calculate the answer.

\#\# Data Description: \{chart\_description\}

\#\# Visualization Code: \{code\}

\#\# Data Path: \{data\_path\}

\#\# Data Format Example: \{data\}

\vspace{\baselineskip}
\#\# Task Type 

Please strictly generate questions according to the following task type requirement:

\{task\}

\vspace{\baselineskip}
\#\# Question Generation Requirements

1. Ensure questions have clear business analysis and practical application value

2. Prioritize generating questions that require multiple calculation steps or statistical analysis

3. Note that question solvers can only see the chart image, not the original chart code and data values

4. While meeting task type requirements, generate appropriately more complex and challenging questions, such as:

- Requiring comprehensive information from multiple dimensions (>3)

- Including multiple steps of reasoning process

- Requiring multiple mathematical operations or complex statistical analysis

- Answers that need in-depth analysis to derive

5. For counting tasks, do not generate questions with answers greater than 20

\vspace{\baselineskip}
\#\# Code Requirements 

1. Use libraries like pandas and numpy for data processing

2. Code must include clear comments explaining the purpose of each step

3. Ensure calculation results are accurate and reliable

4. Only use the provided original data

5. Output necessary intermediate calculation results

6. Code style should be standardized with meaningful variable names

7. For multiple-choice questions, only provide the answer, no need to judge which option is correct

\vspace{\baselineskip}
\#\# Question Types 

1. Multiple-choice: Question includes ABCD four options, answer is a single uppercase letter (A/B/C/D), other options must be incorrect

2. True/False: Question is in interrogative form, answer is Yes or No

3. Fill-in-the-blank: Question is in interrogative or fill-in-the-blank form, answer is a specific number, word, or phrase

4. Short-answer: Question is in interrogative form, answer is a complete sentence not exceeding 50 words

\vspace{\baselineskip}
\#\# Output Format 

```thinking

First provide thinking process, such as explaining what analysis angles and questions can be generated for this task type requirement based on the chart

```

\vspace{\baselineskip}
```json

\{
    "task\_type": "Task type",
    "question\_type": "Question type",
    "question": "Question text",
    "options": "Option text (string, empty for non-multiple-choice questions)"
\}

```

\vspace{\baselineskip}
```python

\# Import required libraries

import pandas as pd

import numpy as np

\# Loading Data from csv file

data\_file\_path = "{data\_path}"

df = pd.read\_csv(data\_file\_path)

\# Data processing and calculation code

...

\# Print intermediate results

print("Average of metric a:", average\_a)

...

\# Print final results

print("Final result:", result)

```

\end{promptblock}
    \caption{ Prompt template for the first stage in Q\&A generation.}
  \label{fig:qa_prompt_1}
\end{figure*}

\begin{figure*}
\begin{promptblock}[LLM Prompt]{lightblue}{prompt:qagen2}

The code execution result is:

\{code\_output\}

\vspace{\baselineskip}
Please use this as data support to provide detailed reasoning analysis for the question and generate the final answer.

Specifically, for multiple-choice questions, if you believe all options are incorrect or multiple options are correct, please modify the options to ensure: the final answer is completely correct, and all other options except the answer are incorrect.

\vspace{\baselineskip}
\#\# Generation Requirements 

1. Please fully trust the correctness of code execution results.

2. All reasoning processes should be expressed as analysis and calculation of visual information from the chart. Don't mention that you referenced code or output results; instead, present them as if they were results you calculated yourself based on visual chart information.

3. Provide necessary reasoning steps without omitting similar processes. Calculation processes should include formulas and answers.

4. All reasoning processes should be fluent and use concise descriptions without verbosity.

5. Finally, provide a concise and clear answer that meets the answer format requirements for the question type.

6. No code language snippets or color coding should appear.

\vspace{\baselineskip}
\#\# Output Format

```json

\{
    "task\_type": "Task type",
    "question\_type": "Question type",
    "question": "Question text",
    "options": "Option text",
    "explanation": "Detailed step-by-step reasoning process",
    "answer": "Final answer"
\}

```

\vspace{\baselineskip}
\#\# Example Start

\{qa\_example\}

\#\# Example End

\end{promptblock}
    \caption{ Prompt template for the second stage in Q\&A generation.}
  \label{fig:qa_prompt_2}
\end{figure*}

\begin{figure*}
\begin{promptblock}[Judge Prompt]{lightblue}{prompt:judge}
Compare the ground truth with the prediction from AI model and determine if the prediction is correct. The question is about an image, which we have not given here. You need to determine whether the model's prediction is consistent with the ground truth. No points will be awarded for wrong answers, over answers or under answers. The reasoning process in the prediction does not need to be considered too much, you only need to determine if the final answer is consistent. There are times when the answer may have a different form of expression and some variation is acceptable. 

Notice:

1. The provided ground truth is absolutely correct and should be fully trusted.

2. Different expressions of units are acceptable. (e.g., "5" vs "5 meters" and "5" vs "5 million" are equivalent if they refer to the same measurement)

3. Numbers with/without "\%" are equivalent (e.g., "5\%" vs "5" are equivalent)

4. After removing units or "\%", if both prediction and ground truth are numbers, an error margin within 5\% error is acceptable. 

5. If the ground truth is provided as multiple arrays, prediction matching any one of them will be considered correct.

6. When the question asks about years:  The prediction must match exactly with the ground truth.








\vspace{\baselineskip}
\#\# Question: \{question\}

\#\# Ground Truth: \{answer\}

\#\# Prediction: \{prediction\}

Now, let's take a analysis and then provide your judgement. Your response must follow the format below:

Analysis: (analyze the correctness briefly) 

Correctness: (Yes or No)

\end{promptblock}
    \caption{ Prompt template for LLM judge model.}
  \label{fig:judge_prompt}
\end{figure*}

\end{document}